\documentclass[preprint,12pt]{elsarticle} 

\usepackage[a4paper, margin=2.5cm]{geometry}

\usepackage{natbib}
\usepackage[utf8]{inputenc} 
\usepackage[T1]{fontenc}    
\usepackage[colorlinks=true, linkcolor=black, urlcolor=blue, citecolor=black, pdfborder={0 0 0}, hypertexnames=false]{hyperref}
\usepackage{url}            
\usepackage{booktabs}       
\usepackage{nicefrac}       
\usepackage{lipsum,graphicx,ulem}
\usepackage{multirow}%
\usepackage{amsmath,amssymb,amsfonts}%
\usepackage{amsthm}%
\usepackage{mathrsfs}%
\usepackage[title]{appendix}%
\usepackage{xcolor}%
\usepackage{textcomp}%
\usepackage{manyfoot}%
\usepackage{algorithm}%
\usepackage{algorithmicx}%
\usepackage{algpseudocode}%
\usepackage{listings, placeins}%
\usepackage[labelfont=bf, labelsep=period]{caption}
\usepackage{subcaption} %

\graphicspath{{figure/}}

\theoremstyle{thmstyleone}%
\theoremstyle{thmstyletwo}%

\theoremstyle{thmstylethree}%

\usepackage{bm}
\usepackage{comment}
\usepackage{tikz}

\newtheorem{prop}{Proposition}
\newcommand{\R}{\mathbb{R}}
\newcommand{\X}{\mathcal{X}}
\newcommand{\Z}{\mathcal{Z}}
\newcommand{\Loss}{\mathcal{L}}
\newcommand\eqdef{\overset{\mathrm{def}}{=}}

\newcommand\vect{\mathrm{vec}}

\DeclareMathOperator*{\argmin}{argmin}

\def\tilde{\widetilde}

\begin{document}

\begin{frontmatter}
\title{TensorProjection Layer: A Tensor-Based Dimension Reduction Method in Deep Neural Networks}

\author[1]{Toshinari Morimoto\corref{cor1}}\ead{d09221002@ntu.edu.tw}
\author[2]{Su-Yun Huang}\ead{syhuang@stat.sinica.edu.tw}

\cortext[cor1]{Corresponding author}

\address[1]{Department of Mathematics, National Taiwan University, No. 1, Section 4, Roosevelt Rd, Da'an District, Taipei, 10617, Taiwan}
\address[2]{Institute of Statistical Science, Academia Sinica, No. 128, Section 2, Academia Rd, Nangang District, Taipei City, 115, Taiwan}


\begin{abstract}
In this paper, we propose a dimension reduction method specifically designed for tensor-structured feature data in deep neural networks.
The method is implemented as a hidden layer, called the TensorProjection layer, which transforms input tensors into output tensors with reduced dimensions through mode-wise projections. 
The projection directions are treated as model parameters of the layer and are optimized during model training. 
Our method can serve as an alternative to pooling layers for summarizing image data, or to convolutional layers as a technique for reducing the number of channels.
We conduct experiments on tasks such as medical image classification and segmentation, integrating the TensorProjection layer into commonly used baseline architectures to evaluate its effectiveness. 
Numerical experiments indicate that the proposed method can outperform traditional downsampling methods, such as pooling layers, in our tasks, suggesting it as a promising alternative for feature summarization.
\end{abstract}


\end{frontmatter}

%
%

\section{Introduction}\label{sec:introduction}

In convolutional neural networks, pooling layers are widely used to summarize the feature data extracted from convolutional layers, reducing unnecessary noise and improving robustness to translation \cite{Dominik2010}.
Additionally, by reducing the dimension of the feature data, pooling layers help decrease the number of parameters in fully connected layers, contributing to more stable learning and preventing overfitting.
This article introduces an alternative dimension reduction method, called the TensorProjection layer, which can serve as a potential replacement for pooling layers or other downsampling techniques.

To highlight the features of the TensorProjection layer, we first review related works by other authors.
HOPE, proposed by \citet{Zhang2016}, is a framework that combines dimension reduction with latent variable modeling. 
It projects high-dimensional data into a lower-dimensional space where a probabilistic model is applied.
\citet{Pan2017} extended the application of HOPE to the training of convolutional neural networks.  
In their work, the feature map produced by a convolutional layer is decomposed into local regions, vectorized, and then projected onto a lower-dimensional space via an orthogonal matrix.
Specifically, the projection is given by $\bm{z}_i = U^\top \bm{x}_i$, where $\bm{x}_i \in \R^{p}$ is the vectorized representation of a local region in the feature map, $\bm{z}_i \in \R^{q}$ is the projected vector, and $U^\top U = I_q$.
Since neural network parameters are typically optimized using gradient descent, the orthogonality constraint $U^\top U = I_q$ may be violated during training.  
To enforce this orthogonal constraint, a penalty term $-\beta D(U)$, where $\beta > 0$ is a tuning parameter, is added to the loss function. 
The penalty term is defined as:
\[
D(U) \eqdef \sum_{i=1}^{q} \sum_{j=i+1}^{q} \frac{|\bm{u}_i^\top \bm{u}_j|}{\|\bm{u}_i\|_2 \cdot \|\bm{u}_j\|_2}, \quad \text{where} \ U \eqdef [\bm{u}_1, \bm{u}_2, \ldots, \bm{u}_q].
\]
This formulation penalizes deviations from orthogonality by minimizing the dot products between different column vectors of $U$.
Our TensorProjection layer differs from previous methods in two key aspects:
\begin{itemize}
\item Unlike prior approaches, our method processes the data directly in its tensor form, without splitting it into local regions or vectorizing the data.
The TensorProjection layer utilizes the mode product of a tensor for projection, preserving the full tensor structure as it passes through the layer.
\item In our approach, the orthogonality of the matrices is strictly enforced throughout the training process, rather than relying on a penalty term to introduce orthogonality.
\end{itemize}

In addition to these works, another relevant approach is Multilinear Principal Component Analysis (MPCA), which is an extension of PCA from vector-structured data to tensor-structured data \cite{Hung2012}.  
Both PCA and MPCA seek projection directions that minimize the reconstruction error.  
Let $\{\X_i\}_{i=1}^n$ be a data set, where each data point $\X_i$ is an order-$m$ tensor of size $p_1 \times \ldots \times p_m$.  
Given a pre-specified dimension $q_1 \times \ldots \times q_m$, consider the mapping of the centered data to a lower-dimensional space:
\[
\Z_i \eqdef (\X_i - \overline{\X}) \times_1 U_1^\top \times_2 U_2^\top \ldots \times_m U_m^\top,
\]
where $\overline{\X} = \frac{1}{n} \sum_{i=1}^n \X_i$, and $U_k \ (k=1,\dots,m)$ are orthogonal matrices of size $p_k \times q_k$ with $q_k < p_k$.  
The reconstructed data are given by:
\[
\tilde{\X}_i \eqdef \Z_i \times_1 U_1 \times_2 U_2 \ldots \times_m U_m + \overline{\X}.
\]
MPCA aims to find the optimal $\{U_k\}_{k=1}^m$ that minimize the reconstruction error:
\[
\argmin_{\{U_k:\, U_k^\top U_k=I_{q_k},k=1,\dots,m\}}~ \sum_{i=1}^n \|\X_i - \tilde{\X}_i\|_F^2,
\]
where $\| \cdot \|_F$ denotes the Frobenius norm for tensors.
While MPCA seeks projection directions that minimize reconstruction error, the TensorProjection layer, in contrast, treats the projection directions $U_1, \ldots, U_m$ as parameters of the underlying neural network, which are learned by minimizing the loss function.

Finally, the TensorProjection layer can be characterized as follows:
\begin{itemize}
\item
{\bf Preserving tensor structure}:
A key feature of the TensorProjection layer is that it reduces the dimension of feature data while preserving the tensor structure.
Flattening the tensor into a vector before projection causes the data to become high-dimensional and results in large projection matrices.
Therefore, the TensorProjection layer performs the projection while preserving the tensor structure.
This keeps the dimension of each mode smaller, significantly reducing the number of parameters to be learned, while preserving the important relationships within the tensor.

\item
{\bf Supervised dimension reduction}:
Another key feature is that, while MPCA determines projection directions by minimizing reconstruction error, the TensorProjection layer obtains the projection directions by optimizing a supervised learning criterion, such as cross-entropy loss.
This allows the orthogonal matrices $\{U_k\}_{k=1}^m$ to incorporate label information. 
Although the TensorProjection layer can also be applied to unsupervised tasks, such as autoencoders, this paper primarily focuses on its use in supervised learning. 
Additionally, in deep neural network training, data is typically processed in batches to reduce memory consumption. 
If MPCA is applied in this setting, projection directions are computed based on small-batch samples, potentially leading to inconsistent projection directions across batches. 
Therefore, it is advantageous to treat the projection directions as trainable parameters optimized across the entire training process, rather than recalculating them for each batch.

\item
{\bf Flexible adjustment for the output dimension}:
Finally, the TensorProjection layer can reduce the dimension to any desired size.
In standard MaxPooling, the output size is determined by the pool size and stride, but the TensorProjection layer is free from these constraints, allowing it to project data to any dimensions.
Moreover, while MaxPooling and Global Average Pooling only reduce the spatial dimensions (height and width) or the channel dimension, the TensorProjection layer can reduce all dimensions simultaneously.
This provides greater flexibility in model design and enables optimal dimension reduction tailored to specific tasks.

\end{itemize}

%
%

\section{Method: TensorProjection layer}\label{sec:TensorProjectionLayer}

%
%
\subsection{Overview of the TensorProjection layer}\label{sec:overview_TPL}

We propose a novel dimension reduction technique by introducing a hidden layer, referred to as the TensorProjection layer, within deep neural networks. 
Similar to pooling layers, this layer is placed after layers that output high-order tensors, such as convolutional layers. 
However, while it is typically positioned after convolutional layers, the TensorProjection Layer can be applied to any layer that outputs high-order tensors, functioning as a dimension reduction layer.

Here, we briefly describe the core concept, with further technical details provided in Sections~\ref{sec:forward_propagation} and~\ref{sec:backprop}.
Let $\{\X_i\}_{i=1}^n$ represent the feature tensors that serve as inputs to the TensorProjection layer, each of size $p_1 \times p_2 \times p_3$, which are the output tensors from a previous layer that produces high-order features.
While the TensorProjection layer can be applied to tensors of any order, we focus on third-order tensors in this paper for simplicity, as our data examples are well-represented in this form.
The TensorProjection layer performs a multilinear mapping, referred to as a mode-$k$ product, or simply mode product, using truncated orthogonal matrices $\{U_k\}_{k=1}^3$ (i.e., $U_k^\top U_k = I_{q_k}$), to project the input tensors into a lower-dimensional feature space. 
This results in output tensors $\{\Z_i\}_{i=1}^n$ of size $q_1 \times q_2 \times q_3$. These orthogonal matrices are treated as trainable parameters within the model.

%
%

\subsection{Forward propagation}\label{sec:forward_propagation}

In this section, we explain the forward propagation process of the TensorProjection layer, detailing how it transforms input tensors into lower-dimensional output tensors through a multilinear mapping. 
The TensorProjection layer reduces the dimension of a third-order tensor using the mode-$k$ product, as reviewed in Appendix \ref{appendix:k-mode-product}.
Let $\{\X_i\}_{i=1}^n$ be the input tensors to the TensorProjection layer, each with dimensions $p_1 \times p_2 \times p_3$.
The output tensors from the TensorProjection layer are given by
\begin{equation}\label{eq:forward_propagation}
\Z_i \eqdef \X_i \times_1 U_1^\top \times_2 U_2^\top \times_3 U_3^\top, \quad (i=1,\ldots,n).
\end{equation}
In Equation~(\ref{eq:forward_propagation}), $\{U_k\}_{k=1}^3$ are the trainable parameters of the TensorProjection layer.
Each $U_k$ is an orthogonal matrix with dimensions $p_k \times q_k \ (k=1,2,3)$ with $q_k < p_k$.
 As a result, the dimensions of each output tensor $\Z_i \ (i=1,\ldots,n)$ are $q_1 \times q_2 \times q_3$.

In deep neural networks, model parameters are typically optimized via gradient descent.
However, directly applying gradient descent to train $\{U_k\}_{k=1}^3$ in (\ref{eq:forward_propagation}) would violate the orthogonality constraint of these matrices.
To address this, we introduce matrices $\{W_k\}_{k=1}^3$ with dimensions $p_k \times q_k$ and redefine $\{U_k\}_{k=1}^3$ as:
\begin{equation}\label{eq:Uk}
U_k \eqdef W_k (W_k^\top W_k)^{-1/2},\quad (k=1,2,3).
\end{equation}
This formulation ensures that $\{U_k\}_{k=1}^3$ always lie in Stiefel manifolds ${\rm St}(p_k,q_k) \eqdef \{U \mid U^\top U = I_{q_k}\} $ ($k=1,2,3$) during the training process. 
Thus, we treat $\{W_k\}_{k=1}^3$ as the trainable parameters instead of $\{U_k\}_{k=1}^3$.
However, $\{W_k^\top W_k\}_{k=1}^3$ may not always be non-singular, which can lead to theoretical and practical issues, such as numerical instability or inversion errors. 
To resolve this, we further modify Equation~(\ref{eq:Uk}) as follows:
\begin{equation}\label{eq:Uk_regularized}
U_k \eqdef W_k (W_k^\top W_k + \epsilon_k^2 I_{q_k})^{-1/2}, \quad (k=1,2,3),
\end{equation}
where $\{\epsilon_k^2\}_{k=1}^3$ are small fixed positive constants.

In summary, we consider $\{W_k\}_{k=1}^3$ as matrices with dimensions $p_k \times q_k$, which are optimized using gradient descent.
We define $\{U_k\}_{k=1}^3$ according to (\ref{eq:Uk_regularized}) instead of (\ref{eq:Uk}).
For each input tensor $\{\X_i\}_{i=1}^n$, we compute the output tensors $\{\Z_i\}_{i=1}^n$ using Equation (\ref{eq:forward_propagation}).

%
%

\subsection{Backpropagation}\label{sec:backprop}

After completing forward propagation of inputs through the TensorProjection layer, it is crucial to optimize the network's parameters effectively. 
Neural networks typically possess a large number of parameters, making it difficult to analytically determine the parameters that minimize the loss function. 
Instead, optimization is performed by updating the parameters using gradient methods. 
In this section, we first review the fundamentals of backpropagation, which serves as the backbone for parameter optimization in neural networks, followed by a discussion on the derivation of the gradient of the loss function with respect to the parameters of the TensorProjection layer.

\subsubsection{Backpropagation essentials}

To optimize the parameters using gradient methods, we need to differentiate the loss function with respect to the parameters. 
Backpropagation is an algorithm for computing the gradients of the loss function with respect to the parameters via the chain rule (see Appendix \ref{appendix:chainrule}), starting from the parameters in the layers closest to the output and working backward through the network.

Consider a neural network consisting of $L$ layers. 
Let $n$ be the number of samples or the minibatch size, and $\Loss$ the loss function to be minimized. 
Let $\{\X_i^{(\ell)}\}_{i=1}^n$ be the input tensors to the $\ell$-th layer ($\ell=1,\dots,L$), and $\{\Z_i^{(\ell)} \}_{i=1}^n$ be the corresponding outputs of the $\ell$-th layer. 
Note that the input to the neural network $\X_i$ is identical to the input to the first layer $\X^{(1)}_i$, and the output from the neural network $\Z_i$ is identical to the output from the final layer $\Z^{(L)}_i$.
Furthermore, let $\theta^{(\ell)}$ be the parameter at the $\ell$-th layer, which can be a scalar, vector, or tensor.
The neural network can be represented as a composite function, where the transformations at each layer are:
\begin{eqnarray*}
\X^{(2)}_i = \Z^{(1)}_i &=& f_1(\X^{(1)}_i \mid \theta^{(1)}), \\
\X^{(3)}_i = \Z^{(2)}_i &=& f_2(\X^{(2)}_i \mid \theta^{(2)}), \\
& \ \vdots &\\
\X^{(L)}_i = \Z^{(L-1)}_i &=& f_{L-1}(\X^{(L-1)}_i \mid \theta^{(L-1)}), \\
\Z^{(L)}_i &=& f_{L} (\X^{(L)}_i \mid \theta^{(L)}).
\end{eqnarray*}
Note that $\X^{(\ell)}_i$ and $\Z^{(\ell)}_i$ are written in calligraphic font for consistency, although the actual inputs and outputs may not necessarily be high-order tensors.

To compute the necessary gradients for optimization, we evaluate $\frac{\partial \Loss}{\partial \vect(\theta^{(\ell)})}$ for each layer $\ell \in \{1,\ldots,L\}$. 
Here, $\vect(\theta^{(\ell)})$ denotes the reshaped form of $\theta$, where a tensor is vectorized (see Appendix \ref{appendix:vec_operation}). 
Assuming that $\frac{\partial \Loss}{\partial \vect(\Z_i^{(\ell)})^\top}, \frac{\partial \Loss}{\partial \vect(\Z_i^{(\ell+1)})^\top}, \ldots, \frac{\partial \Loss}{\partial \vect(\Z_i^{(L)})^\top}$ have already been computed for each $i=1,\ldots,n$, we can compute the gradient with respect to $\theta^{(\ell)}$ using the chain rule:
\begin{eqnarray*}
\frac{\partial \Loss}{\partial \vect(\theta^{(\ell)})^\top} &=& \frac{\partial \Loss}{\partial \vect(\Z^{(\ell)})^\top} \cdot \frac{\partial \vect(\Z^{(\ell)})}{\partial \vect(\theta^{(\ell)})^\top} 
= \sum_{i=1}^n \frac{\partial \Loss}{\partial \vect(\Z_i^{(\ell)})^\top} \cdot \frac{\partial \vect(\Z_i^{(\ell)})}{\partial \vect(\theta^{(\ell)})^\top},
\end{eqnarray*}
where $\Z^{(\ell)}$ is a tensor formed by concatenating $\{\Z^{(\ell)}_i\}_{i=1}^n$.
Notice that $\frac{\partial \vect(\Z_i^{(\ell)})}{\partial \vect(\theta^{(\ell)})^\top}$ can be computed by focusing only on the $\ell$-th layer.

Thus, the derivative of the loss function with respect to the parameters in the $\ell$-th layer can be computed. 
However, to extend the calculation to the $(\ell-1)$-th layer, we need to compute $\frac{\partial \Loss}{\partial \vect(\Z_i^{(\ell-1)})^\top}$ in a similar manner. 
This can also be computed sequentially using the chain rule as follows:
\begin{eqnarray*}
\frac{\partial \Loss}{\partial \vect(\Z_i^{(\ell-1)})^\top} &=& \frac{\partial \Loss}{\partial \vect(\X_i^{(\ell)})^\top} 
=\frac{\partial \Loss}{\partial \vect(\Z_i^{(\ell)})^\top} \cdot \frac{\partial \vect(\Z_i^{(\ell)})}{\partial \vect(\X_i^{(\ell)})^\top},
\end{eqnarray*}
where $\frac{\partial \vect(\Z_i^{(\ell)})}{\partial \vect(\X_i^{(\ell)})^\top}$ is computed by focusing only on the $\ell$-th layer.
By following these steps, we can efficiently compute the gradients necessary for optimizing the neural network parameters via backpropagation.

\subsubsection{Gradients with respect to parameters in TensorProjection layer}

Recall that the key trainable parameters in the TensorProjection layer are $\{W_k\}_{k=1}^3$.
To perform gradient descent for optimization, we need to compute both $\{\frac{\partial \Loss}{\partial \X_i}\}_{i=1}^n$ and $\{\frac{\partial \Loss}{\partial W_k } \}_{k=1}^3$. 
The gradients $\{\frac{\partial \Loss}{\partial W_k } \}_{k=1}^3$ are essential for updating the parameters of the TensorProjection layer, while 
$\{\frac{\partial \Loss}{\partial X_i} \}_{i=1}^n$ are required to propagate the gradients back to the preceding layers. 
For convenience, we further introduce the following matrices:
\[
M_k \eqdef W_k^\top W_k + \epsilon_k^2 I_{q_k} \quad {\rm and}\quad
G_k \eqdef M_k^{-1/2} = \left(W_k^\top W_k + \epsilon_k^2 I_{q_k}\right)^{-1/2} , \quad (k=1,2,3).
\]
With this, we express $U_k$ as $U_k = W_k G_k$.
The gradients related to the TensorProjection layer, which are crucial for backpropagation, are presented in Proposition~\ref{prop:gradients_of_TensorProjectionLayer}. 

%
%

\begin{prop}[Gradients related to the TensorProjection layer]\label{prop:gradients_of_TensorProjectionLayer}
Let $\{\X_i\}_{i=1}^n$ be the input data to the TensorProjection layer, and let $\{\Z_i\}_{i=1}^n$ be the output as defined by Equation~(\ref{eq:forward_propagation}). 
The gradients required for backpropagation are calculated as follows:
\begin{eqnarray}
\frac{\partial \Loss}{\partial \vect(\X_i)^\top} 
&=& \frac{\partial \Loss}{\partial \vect(\Z_i)^\top} \frac{\partial \vect(\Z_i)}{\partial \vect(\X_i)^\top}  = \frac{\partial \Loss}{\partial \vect(\Z_i)^\top} \left(U_3^\top \otimes U_2^\top \otimes U_1^\top\right),  \label{eq:dLdvecXi} \\
\frac{\partial \Loss}{\partial \vect(W_k)^\top} &=& \sum_{i=1}^n \frac{\partial \Loss}{\partial \vect(\Z_i)^\top} \frac{\partial \vect(\Z_i)}{\partial \vect(U_k)^\top} \frac{\partial \vect(U_k)}{\partial \vect(W_k)^\top}.  \label{eq:dLdvecWk}
\end{eqnarray}
In Equation (\ref{eq:dLdvecWk}), the two remaining derivatives, $\frac{\partial \vect(\Z_i)}{\partial \vect(U_k)^\top}$ and $\frac{\partial \vect(U_k)}{\partial \vect(W_k)^\top}$, are provided below. 

First, for the derivative of $\vect(\Z_i)$ with respect to $\vect(U_k)$, we have:
\begin{eqnarray}
\frac{\partial \vect(\Z_i)}{\partial \vect(U_1)^\top} &=& P^{(1)}_{q_1,q_2,q_3} \left[ \left\{ \left(U_3^\top \otimes U_2^\top \right) \X_{i_{(1)}}^\top \right\} \otimes I_{q_1} \right] K_{p_1, q_1}, \label{eq:dvecZidvecU1} \\
\frac{\partial \vect(\Z_i)}{\partial \vect(U_2)^\top} &=& P^{(2)}_{q_1,q_2,q_3} \left[ \left\{ \left(U_3^\top \otimes U_1^\top \right) \X_{i_{(2)}}^\top \right\} \otimes I_{q_2} \right] K_{p_2, q_2},\label{eq:dvecZidvecU2} \\
\frac{\partial \vect(\Z_i)}{\partial \vect(U_3)^\top} &=& P^{(3)}_{q_1,q_2,q_3} \left[ \left\{ \left(U_2^\top \otimes U_1^\top \right) \X_{i_{(3)}}^\top \right\} \otimes I_{q_3} \right] K_{p_3, q_3}, \label{eq:dvecZidvecU3} 
\end{eqnarray}
where $P^{(k)}_{q_1,q_2,q_3}$ are permutation matrices determined by $q_1,q_2, q_3$ and $k$, satisfying $\vect(\Z_i) = P_{q_1,q_2,q_3}^{(k)} \vect(\Z_{i_{(k)}})$, and $K_{p_k,q_k}$ are commutation matrices. 
The terms $\X_{i_{(k)}}$ and $\Z_{i_{(k)}}$ represent the matrices obtained by unfolding (see Appendix \ref{appendix:unfolding_folding}) the third-order tensors $\X_i$ and $\Z_i$ along the $k$th mode.

Second, the derivative of $\vect(U_k)$ with respect to $\vect(W_k)$ is computed as follows:
\begin{equation}
\frac{\partial \vect(U_k)}{\partial \vect(W_k)^\top} = G_k^\top \otimes I_{p_k} + \left(I_{q_k} \otimes W_k\right) \frac{\partial \vect(G_k)}{\partial \vect(W_k)^\top}, \label{eq:vecUkvecWk}
\end{equation}
where the derivative of $\vect(G_k)$ with respect to $\vect(W_k)$ is:
\begin{equation}
\frac{\partial \vect(G_k)}{\partial \vect(W_k)^\top} = \frac{\partial \vect(G_k)}{\partial \vect(M_k)^\top} \frac{\partial \vect(M_k)}{\partial \vect(W_k)^\top}. \label{eq:vecGkvecWk}
\end{equation}
The remaining terms, $\frac{\partial \vect(G_k)}{\partial \vect(M_k)^\top}$ and $\frac{\partial \vect(M_k)}{\partial \vect(W_k)^\top}$, are given by:
\begin{eqnarray}
\frac{\partial \vect(G_k)}{\partial \vect(M_k)^\top} &=& -\frac{1}{2} M_k^{-3/4} \otimes M_k^{-3/4} \label{eq:vecGkvecMk} \\
\frac{\partial \vect(M_k)}{\partial \vect(W_k)^\top} &=& K_{q_k, q_k} \left(I_{q_k} \otimes W_k^\top\right) + I_{q_k} \otimes W_k^\top .\label{eq:vecMkvecWk}
\end{eqnarray}
These expressions provide the complete set of gradients necessary for backpropagation through the TensorProjection layer.
The proof for Proposition \ref{prop:gradients_of_TensorProjectionLayer} is provided in Appendix \ref{appendix:proof}.
\end{prop}

%
%

\section{Data examples}\label{sec:examples}

In this section, we examine the performance of the TensorProjection layer by applying it to three different data examples: retinal OCT image classification, chest X-ray image classification, and gastrointestinal polyp image segmentation. 
The results for each data example are presented separately in  Sections~\ref{sec:retinal_oct}, \ref{sec:chest_x_ray} and \ref{sec:kvasir_segmentation}, respectively.

%
%

\subsection{Classification of retinal OCT images}\label{sec:retinal_oct}

%
%

\subsubsection{Dataset and task}

In this section, we classify retinal diseases using convolutional neural networks (CNNs) with optical coherence tomography (OCT) images of the retina. 
The dataset used is from \cite{Kermany2018}, consisting of OCT images categorized into four classes, each corresponding to a retinal condition.
\footnote{The dataset is also publicly available for download at \url{https://www.kaggle.com/datasets/paultimothymooney/kermany2018}.}
The first class, \texttt{NORMAL}, represents healthy retinas, the second class, \texttt{CNV}, refers to lesions associated with choroidal neovascularization, the third class, \texttt{DME}, corresponds to diabetic macular edema, and the fourth class, \texttt{DRUSEN}, indicates the presence of drusen found in early age-related macular degeneration. 
The number of samples in the training and testing datasets for each class is shown in Table \ref{table:oct_class_counts}.
 
\begin{table}[h!]
\centering
\small
\begin{tabular}{lccc}
\hline
Class   & Train & Validation& Percentage (\%) \\
\hline
NORMAL  & 35973 & 15417      & 47.01  \\
CNV     & 26218 & 11237      & 34.27  \\
DME     & 8118  & 3480       & 10.61  \\
DRUSEN  & 6206  & 2660       & 8.11   \\
\hline
\end{tabular}
\caption{Retinal OCT images: number of images in training and test sets with percentages.}
\label{table:oct_class_counts}
\end{table}

%
%

\subsubsection{Models}

In this section, we describe the deep neural network models used in our experiments. 
The architectures of the models are shown in Table \ref{table:oct_models}. 
And, each input data is a $96\times 96\times 3$ color image.
Below, we describe the three models:

\begin{itemize}
\item {\bf Baseline model:} The baseline model adopts a standard CNN architecture composed of four convolutional layers with ReLU activation followed by MaxPooling layers. 
The model eventually connects to fully connected dense layers. 
To ensure that the spatial dimensions, width and height, remain unchanged through the convolutional layers, we apply \texttt{same} padding.
\item {\bf The 1st TensorProjection Layer model:} 
    The first variant model, namely the 1st TPL model, replaces the final MaxPooling layer of the baseline model with a TensorProjection layer of the same dimensions. 
    By unifying the output dimensions, we aim to allow a more comparable evaluation of the models. 
    The reason for only replacing the final MaxPooling layer is that, in earlier layers closer to the input, neighboring pixel values are expected to be more similar, making dimension reduction through pooling more effective. 
    However, as we approach the output layers and the dimensions of the data gradually decrease, pooling may excessively reduce critical information. 
    Therefore, we expect that dimension reduction through the TensorProjection layer will be more beneficial in this context.
\item {\bf The 2nd TensorProjection Layer model:} The second variant model, namely the 2nd TPL model, also incorporates a TensorProjection layer. 
    However, unlike the first variant, this model performs dimension reduction along the channel axis as well. 
    This allows us to test the TensorProjection layer's ability to reduce dimension across multiple axes while preserving the tensor structure.
\end{itemize}

\begin{table}[h!]
    \centering
    \small
    \begin{minipage}{\textwidth}
        \centering
        \begin{tabular}{|l|l|r|}
        \hline
        Layer (activation)         & Output Shape     & Param \#  \\
        \hline
        Conv2D (relu)               & ($n$, 96, 96, 32)  & 896    \\
        MaxPooling2D         & ($n$, 48, 48, 32)  & 0      \\
        Conv2D (relu)               & ($n$, 48, 48, 32)  & 9248   \\
        MaxPooling2D         & ($n$, 24, 24, 32)  & 0      \\
        Conv2D (relu)               & ($n$, 24, 24, 32)  & 9248   \\
        MaxPooling2D         & ($n$, 12, 12, 32)  & 0      \\
        Conv2D (relu)               & ($n$, 12, 12, 32)  & 9248   \\
        {\bf MaxPooling2D}         & \bf ($\bm n$, 6, 6, 32)    & 0      \\
        Flatten              & ($n$, 1152)        & 0      \\
        Dense (relu)                & ($n$, 256)         & 295168 \\
        Dense (softmax)                & ($n$, 4)           & 1028   \\
        \hline
        \end{tabular}
        \subcaption{Baseline model.} 
    \end{minipage}
    
    \vspace{0.25cm}  
    
    \begin{minipage}{\textwidth}
        \centering
        \begin{tabular}{|l|l|r|}
        \hline
        Layer (activation)         & Output Shape     & Param \#  \\
        \hline
        Conv2D (relu)               & ($n$, 96, 96, 32)  & 896    \\
        MaxPooling2D         & ($n$, 48, 48, 32)  & 0      \\
        Conv2D (relu)               & ($n$, 48, 48, 32)  & 9248   \\
        MaxPooling2D         & ($n$, 24, 24, 32)  & 0      \\
        Conv2D (relu)               & ($n$, 24, 24, 32)  & 9248   \\
        MaxPooling2D         & ($n$, 12, 12, 32)  & 0      \\
        Conv2D (relu)               & ($n$, 12, 12, 32)  & 9248   \\
        {\bf TensorProjection}         & \bf ($\bm n$, 6, 6, 32)    & 144      \\
        Flatten              & ($n$, 1152)        & 0      \\
        Dense (relu)                & ($n$, 256)         & 295168 \\
        Dense (softmax)                & ($n$, 4)           & 1028   \\
        \hline
        \end{tabular}
        \subcaption{The 1st TensorProjection Layer (TPL) model.} 
    \end{minipage}
    
    \vspace{0.25cm}  
    
    \begin{minipage}{\textwidth}
        \centering
        \begin{tabular}{|l|l|r|}
        \hline
        Layer (activation)         & Output Shape     & Param \#  \\
        \hline
        Conv2D (relu)               & ($n$, 96, 96, 32)  & 896    \\
        MaxPooling2D         & ($n$, 48, 48, 32)  & 0      \\
        Conv2D (relu)               & ($n$, 48, 48, 32)  & 9248   \\
        MaxPooling2D         & ($n$, 24, 24, 32)  & 0      \\
        Conv2D (relu)               & ($n$, 24, 24, 32)  & 9248   \\
        MaxPooling2D         & ($n$, 12, 12, 32)  & 0      \\
        Conv2D (relu)               & ($n$, 12, 12, 32)  & 9248   \\
        {\bf TensorProjection}         & \bf ($\bm n$, 6, 6, 20)    & 784   \\
        Flatten              & ($n$, 1152)        & 0      \\
        Dense (relu)                & ($n$, 256)         & 295168 \\
        Dense (softmax)                & ($n$, 4)           & 1028   \\
        \hline
        \end{tabular}
        \subcaption{The 2nd TensorProjection Layer (TPL) model.}  
    \end{minipage}

    \caption{
    Model summaries for the baseline model, the 1st TPL model, and the 2nd TPL model in the Retinal OCT image classification task.
    The baseline model follows a standard architecture consisting of four convolutional layers, each followed by a MaxPooling layer, and is connected to a Dense layer for classification.
    In the 1st TPL model, the final MaxPooling layer of the baseline model has been replaced with a TensorProjection layer, with the output size adjusted to match that of the baseline model.
    In the 2nd TPL model, the final MaxPooling layer of the baseline model has been replaced with a TensorProjection layer, and additional dimension reduction is performed along the channel axis.
    }
    \label{table:oct_models}
\end{table}

\pagebreak

%
%
\subsubsection{Training}
We used cross-entropy as the loss function to optimize the model parameters. 
The models were optimized using the Adam optimizer, with 20 training epochs and a batch size of 300. 
The learning rate was set to its default value of 0.001.
Given the influence of randomness in the optimization of deep neural networks, we repeated the experiment 30 times to ensure stability and reproducibility of the results.
The dataset was pre-divided into training, validation and test sets. 
For model training, we used the training set, while the test and validation sets were combined and used as validation data.

%
%

\subsubsection{Results}

\begin{figure}[h]
    \centering
    \begin{minipage}{0.5\textwidth}
        \centering
        \includegraphics[width=\textwidth]{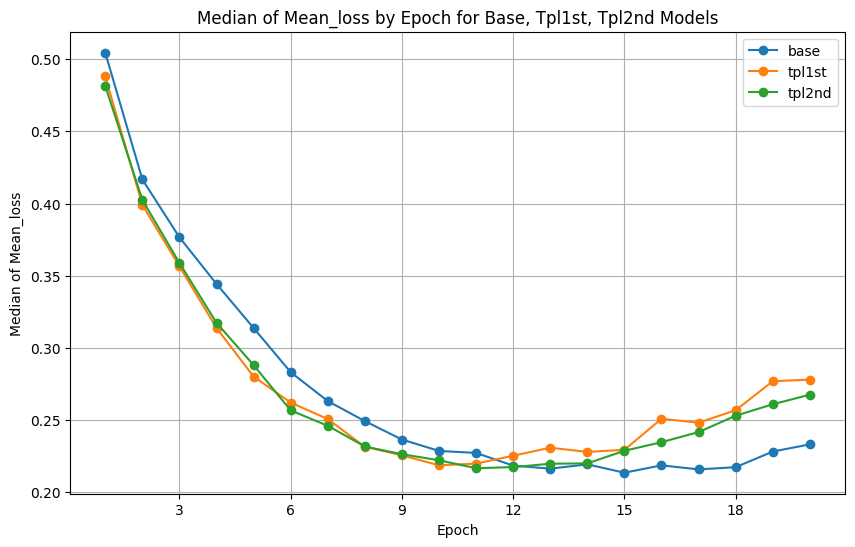}
        \captionsetup{width=1.5\textwidth}
        \caption{Retinal OCT image classification: validation data loss over training epochs. Each point represents the median loss value from 30 repeated runs. The TPL models converge faster than the baseline model, but after epoch 12, the effects of overfitting become more apparent in the TPL models.}
        \label{fig:oct_loss}
    \end{minipage}
    
    \vspace{0.15cm} %
     
    \begin{minipage}{0.5\textwidth}
        \centering
        \includegraphics[width=\textwidth]{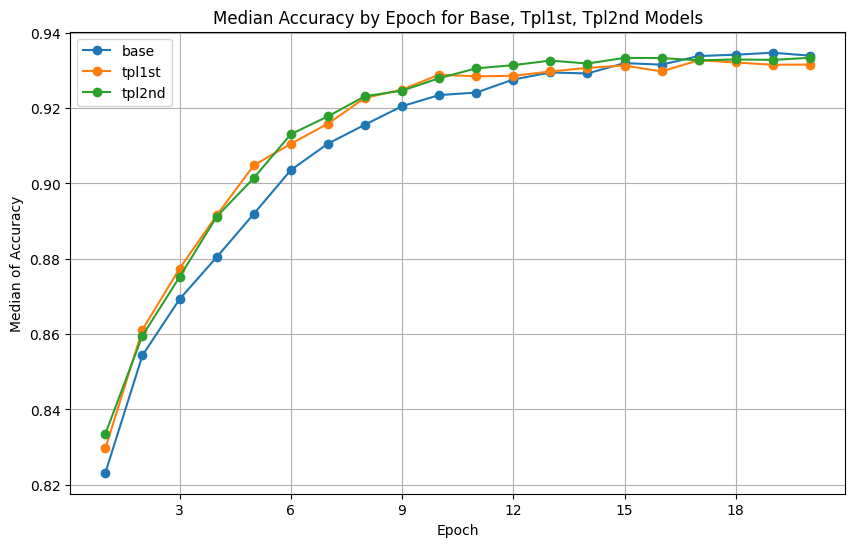}
        \captionsetup{width=1.5\textwidth}
        \caption{Retinal OCT image classification: validation data accuracy over training epochs. The TPL models increase validation accuracy more quickly compared to the baseline model, but eventually, all models reach nearly the same performance level.}
        \label{fig:oct_accuracy}
    \end{minipage}
    
    \vspace{0.15cm} %
    \begin{minipage}{0.5\textwidth}
        \centering
        \includegraphics[width=\textwidth]{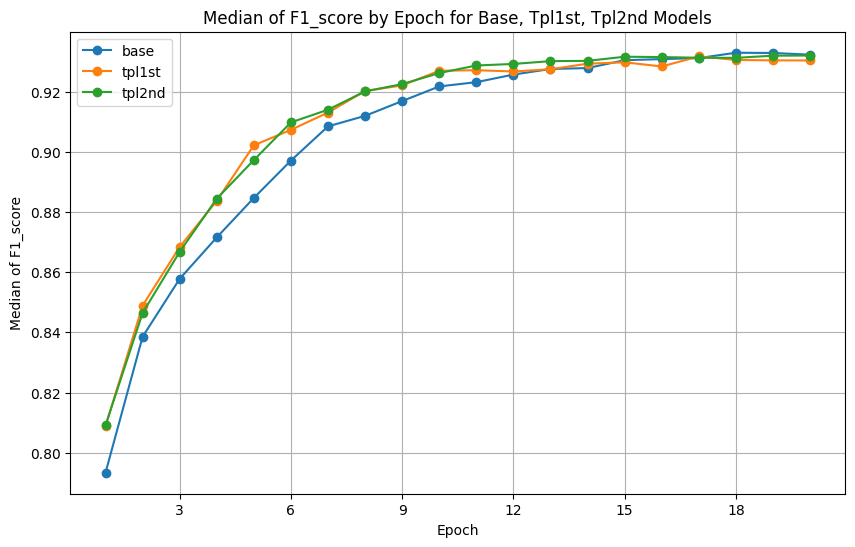}
        \captionsetup{width=1.5\textwidth}
        \caption{Retinal OCT image classification: weighted F1 score for validation data over training epochs. The TPL models improve the F1 score more quickly compared to the baseline model, but eventually, all models converge to nearly the same performance level, as observed with validation accuracy.}
        \label{fig:oct_f1_score}
    \end{minipage}
\end{figure}

Figures \ref{fig:oct_loss}, \ref{fig:oct_accuracy}, and \ref{fig:oct_f1_score} present the median values of the average loss, classification accuracy, and weighted F1 score at each epoch, respectively, computed from 30 independent training trials using the validation data.
In these figures, \texttt{base} refers to the baseline model, \texttt{tpl1st} refers to the first variant model incorporating a TensorProjection layer, and \texttt{tpl2nd} refers to the second variant model.
The following are the observations:
\begin{itemize}
\item Figure \ref{fig:oct_loss} shows that the TPL models (\texttt{tpl1st} and \texttt{tpl2nd}) initially reduce the loss faster than the basline model. 
However, after epoch 12, the effects of overfitting become more pronounced in the TPL models, with their loss values starting to fluctuate and eventually surpassing those of the baseline model. 
\item Similarly, Figures~\ref{fig:oct_accuracy} and~\ref{fig:oct_f1_score} (accuracy and weighted F1 score) show that TPL models reach higher scores earlier compared to the basline model. 
\end{itemize}

These observations suggest that the use of the TensorProjection layer enables efficient feature extraction from high-dimensional data, allowing for performance improvements with fewer epochs. 
Traditional pooling layers (e.g., max pooling) perform dimension reduction by directly selecting the largest value among multiple pixel values. 
This method is highly effective in the early layers where the dimension is large and adjacent feature values are similar.
However, as the model approaches the output layers and the dimension decreases, important features may be lost due to excessive summarization by pooling.

In contrast, the TensorProjection layer performs dimension reduction along projection directions learned through supervised learning, enabling optimal feature summarization based on the data.
However, this parameterized reduction also carries the risk of overfitting. 
As shown in Figure \ref{fig:oct_loss}, the \texttt{tpl1st} model starts to exhibit increasing validation loss midway through training. 
A similar trend is observed in the \texttt{tpl2nd} model, although overfitting is somewhat mitigated due to dimension reduction along the channel axis.
On the other hand, pooling layers, as illustrated in Figure \ref{fig:oct_loss}, show less tendency toward overfitting. 
This is likely due to their non-parametric and simpler approach to dimension reduction.

These results suggest that, by appropriately setting the output dimensions in the TensorProjection layer, efficient feature extraction can be achieved while controlling overfitting to some extent.
In contrast, pooling layers demonstrate superior robustness against overfitting.
That said, in this study, the output dimensions of the TensorProjection layer were intentionally set to match those of the pooling layers for comparison purposes.
In practice, the TensorProjection layer offers flexibility to freely adjust the output dimensions.
Therefore, it does not necessarily carry a significant risk of overfitting.
Practitioners can select the optimal output dimensions to balance data compression efficiency and the risk of overfitting, tailoring the model to suit their needs.

\clearpage

%
%

\subsection{Classification of Chest X-ray images}\label{sec:chest_x_ray}

%
%

\subsubsection{Dataset and task}

In this section, we utilize the COVID-19 Radiography Database, which is publicly accessible on Kaggle \footnote{\url{https://www.kaggle.com/datasets/tawsifurrahman/covid19-radiography-database}}. 
This dataset, developed by a collaborative team of researchers from Qatar University, the University of Dhaka, and several other institutions, offers a comprehensive collection of chest X-ray images categorized into four classes: \texttt{Normal}, \texttt{Lung Opacity}, \texttt{COVID-19}, and \texttt{Viral Pneumonia}.
The distribution of chest X-ray images across different categories in the COVID-19 Radiography Database is presented in Table \ref{table:chest_xray_distribution}.
In accordance with the citation guidelines provided on the Kaggle page, we reference the following key publications: \cite{Chowdhury2020,Rahman2021}.

\begin{table}[h!]
\centering
\small
\begin{tabular}{lcc}
\hline
Class            & Number of Images & Percentage (\%) \\ \hline
Normal           & 10,192           & 48.15\%         \\ 
Lung Opacity     & 6,012            & 28.41\%         \\ 
COVID-19         & 3,616            & 17.08\%         \\ 
Viral Pneumonia  & 1,345            & 6.35\%          \\ \hline
Total            & 21,165           & 100\%           \\
\end{tabular}
\caption{Distribution of chest X-ray images in the COVID-19 Radiography Database}
\label{table:chest_xray_distribution}
\end{table}

%
%

\subsubsection{Models}

DenseNet \cite{Huang2017} is a type of convolutional neural network known for its unique structure called Dense blocks. 
In these blocks, the output of each layer is connected to all subsequent layers via skip connections, which helps mitigate the vanishing gradient problem and enables efficient feature propagation.

In this section, we utilize a pre-trained DenseNet121, excluding its fully connected layers, for feature extraction from images. 
The parameters of DenseNet121 are frozen, and no fine-tuning is performed. The input image size is set to $160\times 160\times 3$, and the output from DenseNet121 is $5\times 5\times 1024$. 
However, directly connecting this output to a fully connected layer would result in an excessive number of parameters, making dimension reduction necessary. 
Therefore, we reduce the number of channels from 1024 to 64 before connecting to the fully connected layer.

The detailed architecture of both the baseline model and the alternative model with a TensorProjection layer is shown in Table~\ref{table:chest_x_ray_models}.
As shown in the table, the baseline model reduces the dimension using a $1\times 1$ convolution layer, applying only linear transformations without any non-linear activations. 
Additionally, in the alternative model, the TensorProjection layer is used to similarly reduce the number of channels to 64.

\begin{table}[h]
    \centering
    \small
    \begin{subtable}{\textwidth}
        \centering
        \begin{tabular}{|l|l|r|}
            \hline
            Layer (type) & Output Shape & Param \# \\
            \hline
            DenseNet121 & $(n, 5, 5, 1024)$ & - \\
            {\bf Conv2D (linear)} & $\bm{(n, 5, 5, 64)}$ & 65,600 \\
            Flatten & $(n, 1600)$ & 0 \\
            Dense (relu) & $(n, 256)$ & 409,856 \\
            Dense (softmax) & $(n, 7)$ & 1,799 \\
            \hline
        \end{tabular}
        \caption{Baseline model summary. }
    \end{subtable}

    \vspace{0.5cm} %

    \begin{subtable}{\textwidth}
        \centering
        \begin{tabular}{|l|l|r|}
            \hline
            Layer (type) & Output Shape & Param \# \\
            \hline
            DenseNet121  & $(n, 5, 5, 1024)$ & - \\
            {\bf TensorProjection} & $\bm{(n, 5, 5, 64)}$ & 65,536 \\
            Flatten & $(n, 1600)$ & 0 \\
            Dense (relu) & $(n, 256)$ & 409,856 \\
            Dense (softmax) & $(n, 7)$ & 1,799 \\
            \hline
        \end{tabular}
        \caption{TPL model summary. }
    \end{subtable}

    \captionsetup{width=\textwidth}
    \caption{Chest X-ray image classification model summaries for the baseline model and the TPL model. 
    The baseline model utilizes DenseNet121 as a feature extractor. Features from DenseNet121 are further reduced along the channel dimension using a Conv2D (linear) layer before being connected to a Dense layer for classification.
    In the TPL model, the Conv2D layer used in the baseline model for channel-wise dimension reduction is replaced by a TensorProjection layer (TPL).
    The output dimensions are adjusted to match those of the baseline model.
    }
    \label{table:chest_x_ray_models}
\end{table}

%
%

\subsubsection{Training}

First, to address class imbalance, we apply data augmentation in advance. 
The augmentation techniques include rotations up to 10 degrees, shifts up to 5\% horizontally and vertically, zooming up to 5\%, and shear transformations within a range of 0.05 radians. 
Flipping is not applied, and any missing areas are filled using the nearest pixel values.
No augmentation is performed on images labeled as \texttt{Normal}, while for images labeled \texttt{Lung Opacity}, one additional augmentation is applied to each original image. 
For \texttt{COVID} images, two augmentations are applied per original image, and for \texttt{Viral Pneumonia} images, seven augmentations are applied per original image.

The training process proceeds as follows:
\begin{itemize}
\item First, the original data (before augmentation) is split into training and validation sets at a ratio of 7:3.
\item Second, the training process uses both the original training images and their corresponding augmented images. The validation process uses only the original images.
\item This training and validation process is repeated 30 times in total. 
\end{itemize}

To reduce computational cost, data augmentation is performed only once in advance. 
This means that the augmentation pattern for each original image remains the same across all experiments. 
However, the data split pattern changes in each experiment, and the training and validation sets are completely separated. 
Additionally, only the original images are used in the validation process.

The model uses cross-entropy as the loss function, and the Adam optimizer is employed for optimization with a batch size of 64. 
However, when using a pre-trained model as in this case, the model tended to converge too quickly, leading to early overfitting. 
To address this, the learning rate was adjusted from the default value to 0.0001.

%
%

\subsubsection{Results}

\begin{figure}[h]
    \centering
    \begin{minipage}{0.5\textwidth}
        \centering
        \includegraphics[width=\textwidth]{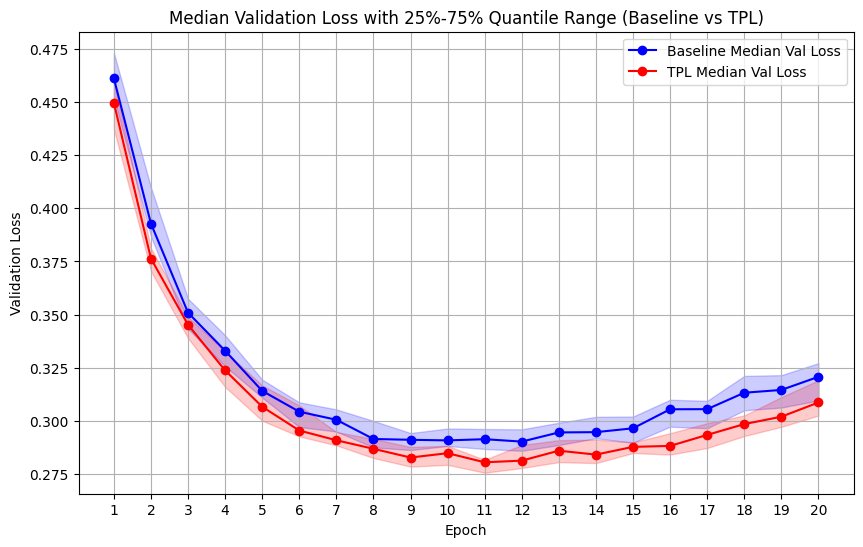}
        \captionsetup{width=1.4\textwidth}
        \caption{ 
        Chest X-ray image classification: median validation loss per epoch for baseline and TPL models, calculated from repeated runs. 
        The shaded range represents the 25th to 75th percentiles. 
        The validation loss for the TPL models remains consistently lower than that of the baseline model throughout the training.
        }
        \label{fig:chest_x_ray_loss}
    \end{minipage}
    
    \vspace{0.15cm} 
    
    \begin{minipage}{0.5\textwidth}
        \centering
        \includegraphics[width=\textwidth]{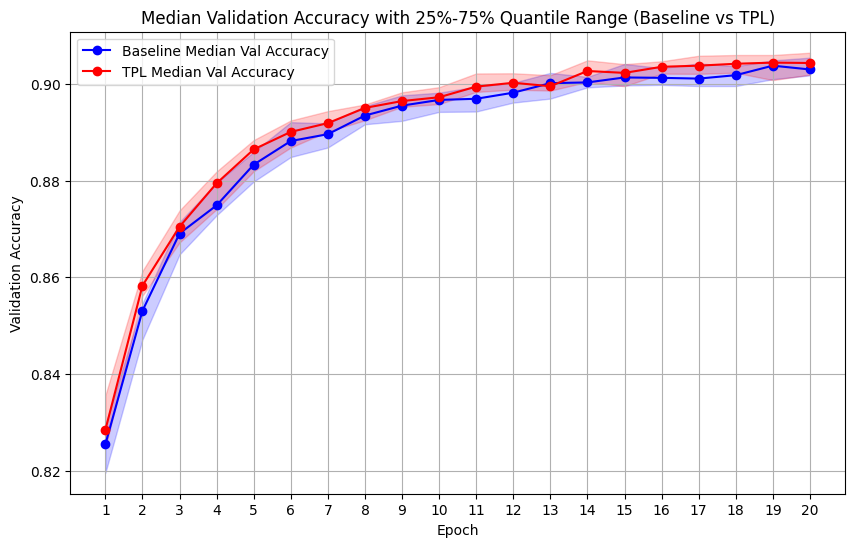}
        \captionsetup{width=1.4\textwidth}
        \caption{
        Chest X-ray image classification: validation median accuracy per epoch for Baseline and TPL models, calculated from repeated runs. 
        The shaded range represents the 25th to 75th percentiles. 
        Throughout the epochs, the TPL models show slightly better performance compared to the baseline model.}
        \label{fig:chest_x_ray_accuracy}
    \end{minipage}
    
    \vspace{0.15cm} 
    
    \begin{minipage}{0.5\textwidth}
        \centering
        \includegraphics[width=\textwidth]{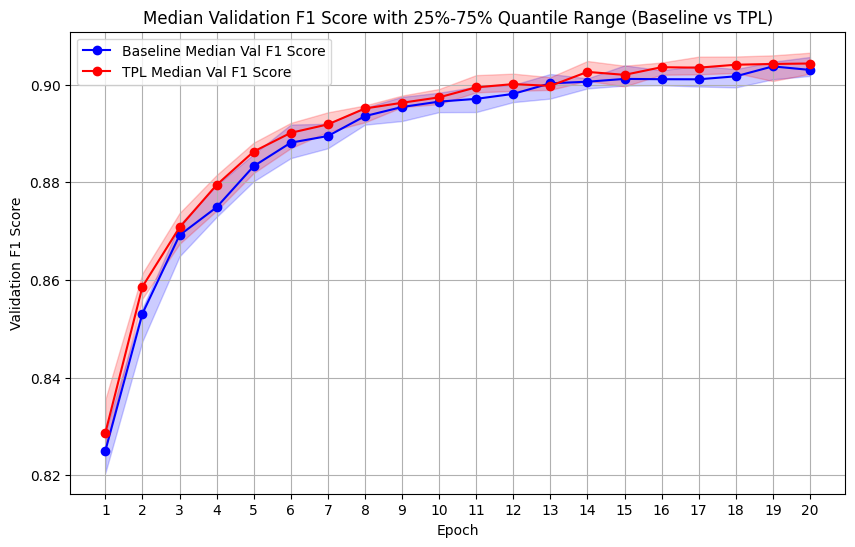}
        \captionsetup{width=1.4\textwidth}
        \caption{
        Chest X-ray image classification: validation median weighted F1score per epoch for Baseline and TPL models, calculated from repeated runs. 
        The shaded range represents the 25th to 75th percentiles. 
        Throughout the epochs, the TPL models show slightly better performance compared to the baseline model.}
        \label{fig:chest_x_ray_f1score}
    \end{minipage}
\end{figure}

Figures \ref{fig:chest_x_ray_loss}, \ref{fig:chest_x_ray_accuracy}, and \ref{fig:chest_x_ray_f1score} display the mean loss, accuracy, and weighted F1 score for the validation data on each epoch. 
In these figures, the median values are computed and plotted based on 30 repeated runs, and the shaded regions represent the 25th to 75th percentile (quantile range) of the results.
The following are the observations:
\begin{itemize}
    \item As shown in Figure \ref{fig:chest_x_ray_loss}, the TPL model consistently maintains lower loss values compared to the baseline model.
    \item Both the baseline and TPL models exhibit an increase in validation loss in the later epochs, indicating signs of overfitting. However, the TPL model consistently keeps the loss lower than the baseline, and the impact of overfitting is relatively smaller.
    \item As shown in Figures \ref{fig:chest_x_ray_accuracy} and \ref{fig:chest_x_ray_f1score}, in both accuracy and F1 score, the TPL model shows slightly better performance than the baseline model throughout all epochs.
\end{itemize}
These results, similar to those presented in the previous section, suggest that the TensorProjection layer performs more efficient dimension reduction along the channel axis than conventional Conv2D.

\clearpage

%
%

\subsection{Segmentation of gastrointestinal polyp images}\label{sec:kvasir_segmentation}

%
%

\subsubsection{Dataset and task}

In this section, we employ the Kvasir-SEG dataset \cite{kvasir2020} for a medical image segmentation task to further investigate the behavior and effectiveness of the TensorProjection layer.
\footnote{The dataset is publicly available for download at \url{https://www.kaggle.com/datasets/debeshjha1/kvasirseg}.}
Kvasir-SEG is a dataset consisting of gastrointestinal polyp images and their corresponding segmentation masks.
Each entry includes an endoscopic image depicting a polyp and a segmentation mask of the same size that indicates the polyp region. 
The dataset includes 1000 polyp images and their corresponding masks (ground truth).

%
%

\subsubsection{Models}

\begin{figure}[h!]
    \centering
    \begin{minipage}{0.95\textwidth}
        \centering
        \includegraphics[width=1\textwidth, height=8cm]{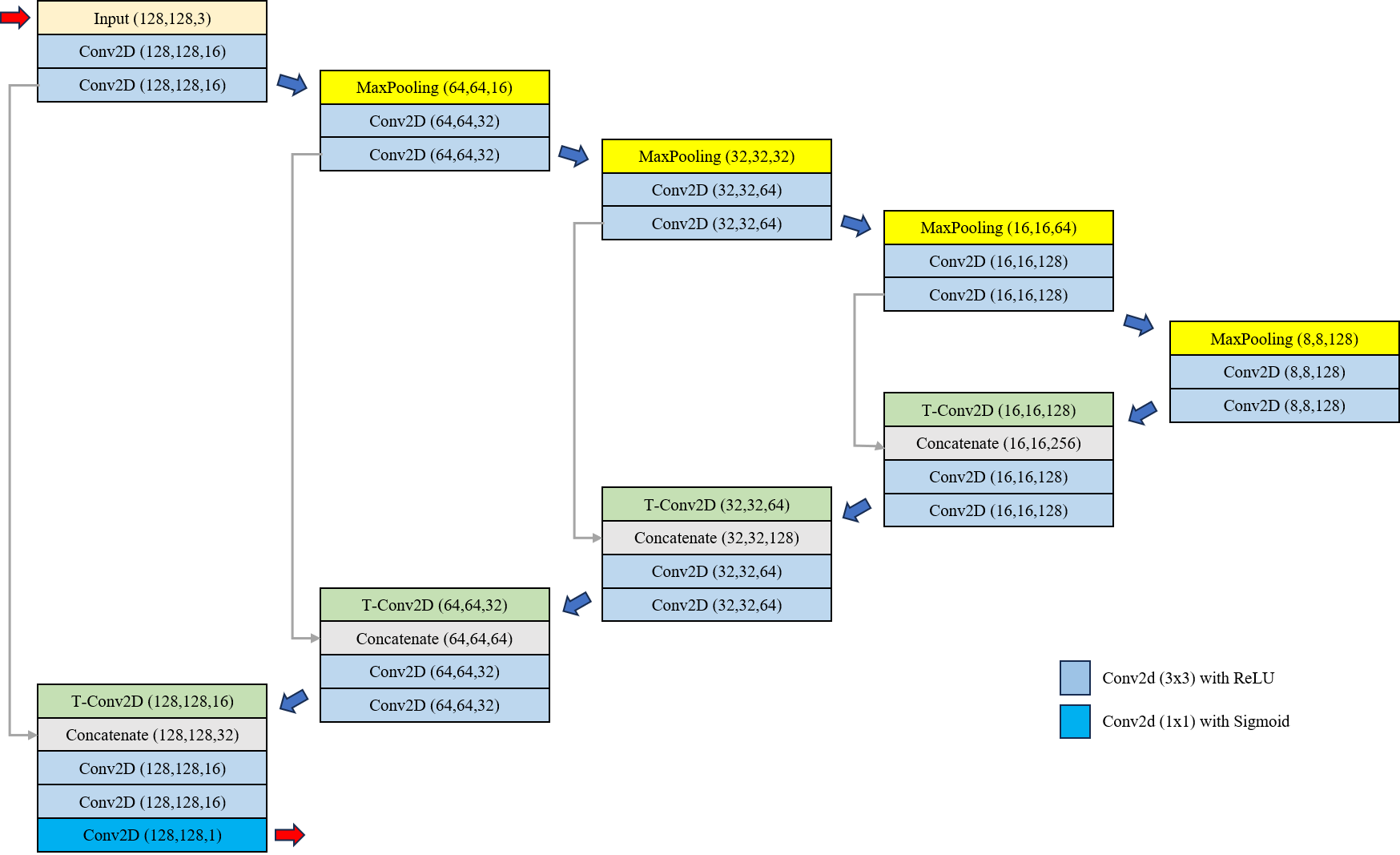}
        \vspace{1mm}
        \caption{The baseline model for polyp image segmentation utilizes a standard U-Net architecture. 
        The encoder consists of four blocks, each with two convolutional layers followed by MaxPooling. 
        In the decoder, four blocks are composed of two convolutional layers followed by upsampling using transpose convolutions, after which features from the corresponding encoder layers are merged via skip connections. 
        Finally, after two additional convolutional layers, the segmentation mask is produced. 
        This figure shows the case where the input size is $128 \times 128$, and the number of channels in the first Conv2D layer is 16. 
        }
        \label{fig:kvasir_baseline_model}
    \end{minipage}
    
    \vspace{0.3cm}

    \begin{minipage}{0.95\textwidth}
        \centering
        \includegraphics[width=1\textwidth, height=8cm]{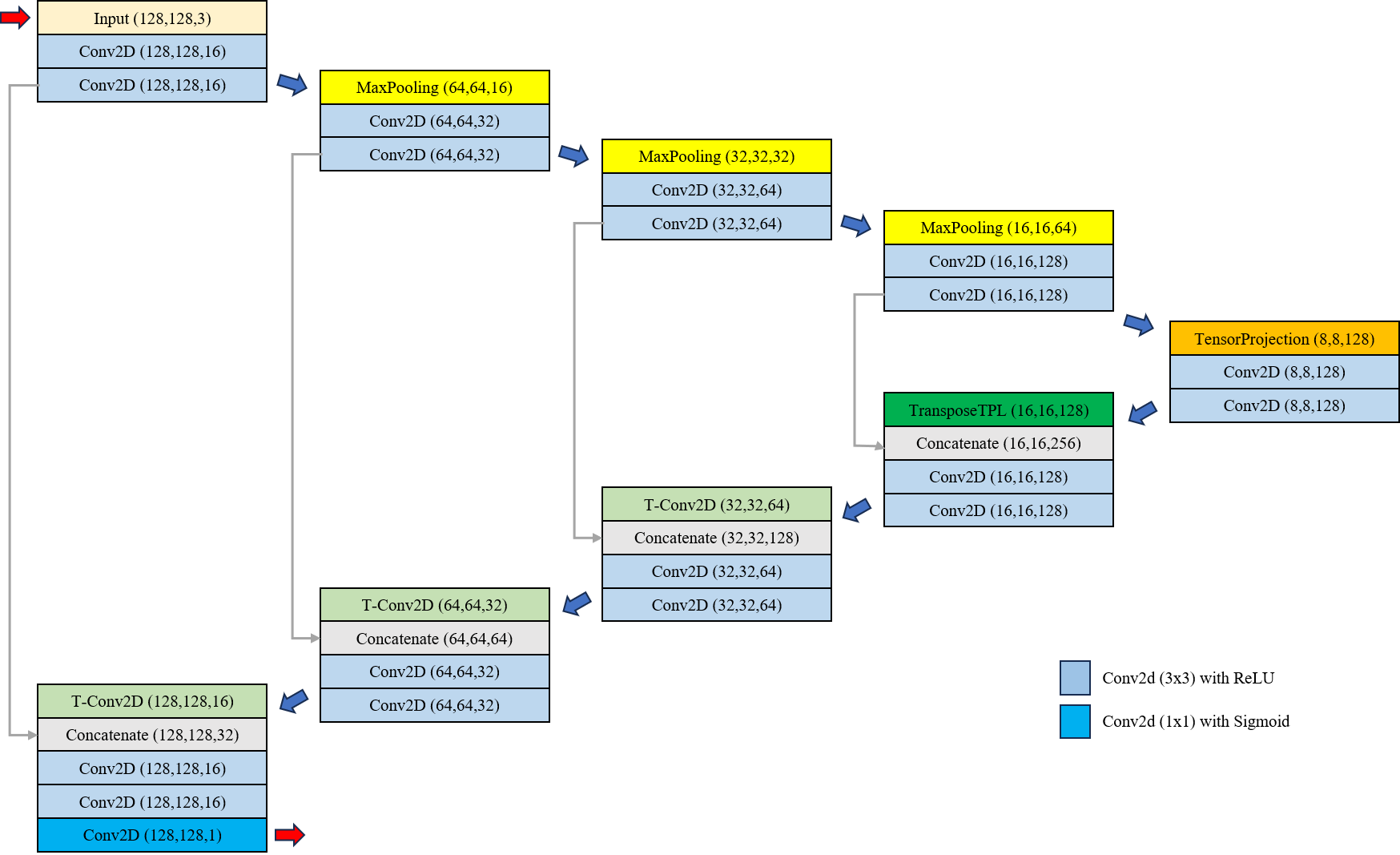}
        \vspace{1mm}
        \caption{The variant model (TPL model) with TensorProjection layers for polyp image segmentation. 
        The final MaxPooling layer (for downsampling) is replaced with a TensorProjection layer, and the first transpose convolutional layer (for upsampling) is replaced with a TransposeTensorProjection layer. 
        This figure shows the case where the input size is $128 \times 128$, and the number of channels in the first Conv2D layer is 16.}
        \label{fig:kvasir_tpl_model}
    \end{minipage}
\end{figure}

Figures \ref{fig:kvasir_baseline_model} and \ref{fig:kvasir_tpl_model} depict the architectures of the models used for this segmentation task. 
The figures show the models with an input size of $128 \times 128$ and 16 channels in the first convolutional layer. 
We also experimented with a configuration where the number of channels in the first convolutional layer was increased to 64. 
In this case, the number of channels in the subsequent layers increased proportionally, becoming four times larger, except for the output layer.
Both models are based on the U-Net architecture \cite{unet}, which consists of an encoder to extract features from input images and a decoder to reconstruct the segmentation mask from the encoded features.
A key feature of U-Net is the merging of feature maps in the encoder with upsampled data features in the decoder, facilitating better reconstruction. 
Additionally, batch normalization is applied after each convolutional layer in the decoder to stabilize feature maps during upsampling and enhance performance.
The baseline model, shown in Figure~\ref{fig:kvasir_baseline_model}, follows the standard U-Net design.
In the variant model (TPL model), shown in Figure~\ref{fig:kvasir_tpl_model}, the final MaxPooling layer in the encoder is replaced with a TensorProjection layer. 
Moreover, the first transpose convolutional layer in the decoder is replaced with a TransposeTensorProjection layer designed specifically for upsampling.
The TransposeTensorProjection layer performs upsampling by transforming the input tensors using trainable projection matrices. The output tensor $\Z_i$ in this layer is defined as: 
\begin{equation}\label{transpose_tpl}
 \Z_i \eqdef \X_i \times_1 U_1 \times_2 U_2 \times_3 U_3 \quad (i=1,\ldots,n),
\end{equation}
where the transformation matrices $U_k$ are given by: 
\[
U_k \eqdef W_k(W_k^\top W_k + \epsilon^2 I_{p_k})^{-1/2} \quad (k=1,2,3),
\]
with $W_k$ being $q_k \times p_k$ matrices, and $p_k < q_k$. 
While the TensorProjection layer is used in the encoder for feature dimension reduction, the TransposeTensorProjection layer in the decoder serves the opposite purpose--upsampling to increase the feature dimension.

Finally, Dice loss, which is defined as 1 minus the Dice coefficient, was employed as the loss function for model optimization.

\pagebreak

%
%

\subsubsection{Training}

The original dataset consists of 1,000 images and their corresponding masks. 
To expand the dataset, we applied data augmentation, generating 20 augmented versions for each original image and mask, resulting in a total of 20,000 images and masks.
The augmentations included the following transformations: rotation (up to 30 degrees), width and height shifts (up to 10\%), shearing (up to 0.05 radian), zooming (up to 5\%), brightness adjustment (between 80\% and 120\% of the original brightness), color shifting (up to 20 units), as well as horizontal and vertical flipping. Nearest-neighbor filling was applied to address any gaps created by the transformations.

Similar to Section \ref{sec:chest_x_ray}, the training proceeds as follows:
\begin{itemize}
    \item First, the original 1,000 images and their corresponding masks are randomly split into 700 for training and 300 for validation.
    \item Second, the models are trained using the pairs of augmented images and their masks, generated from the 700 original training images and corresponding masks. However, the validation set uses only the original 300 images and masks.
    \item To account for randomness in model training, 30 independent experiments are conducted.
\end{itemize}

To reduce computational cost, all augmentations were performed in advance, as in Section \ref{sec:chest_x_ray}.  
However, the data split patterns change in each independent experiment.  
The training data are completely separated from the validation data in each experiment, and the validation data and their corresponding augmented images are not included in the training set.

Dice loss was used as the loss function for parameter optimization, and the parameters were optimized using the Adam algorithm with the default learning rate. The batch size was set to 50.

%
%

\subsubsection{Results}

\begin{figure}[h!]
    \centering
    \begin{minipage}{0.48\textwidth}
        \centering
        \includegraphics[width=\textwidth]{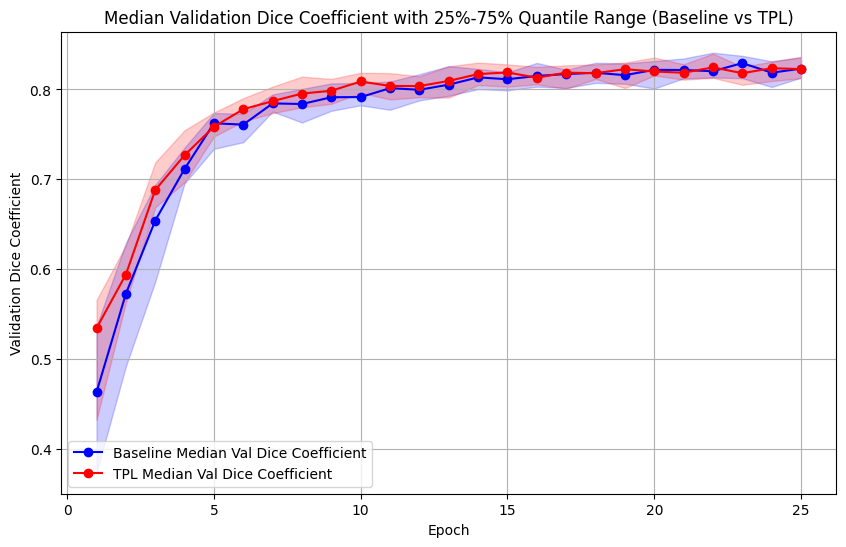}
        \subcaption{Dice score ($128 \times 128$, 16 channels)}
    \end{minipage}
    \hfill
    \begin{minipage}{0.48\textwidth}
        \centering
        \includegraphics[width=\textwidth]{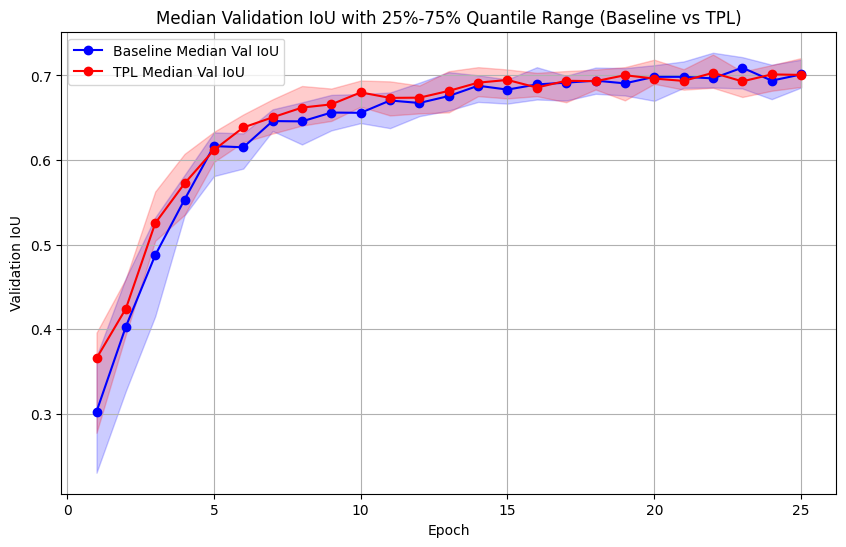}
        \subcaption{IoU score ($128 \times 128$, 16 channels)}
    \end{minipage}

    \vspace{0.15cm}

    \begin{minipage}{0.48\textwidth}
        \centering
        \includegraphics[width=\textwidth]{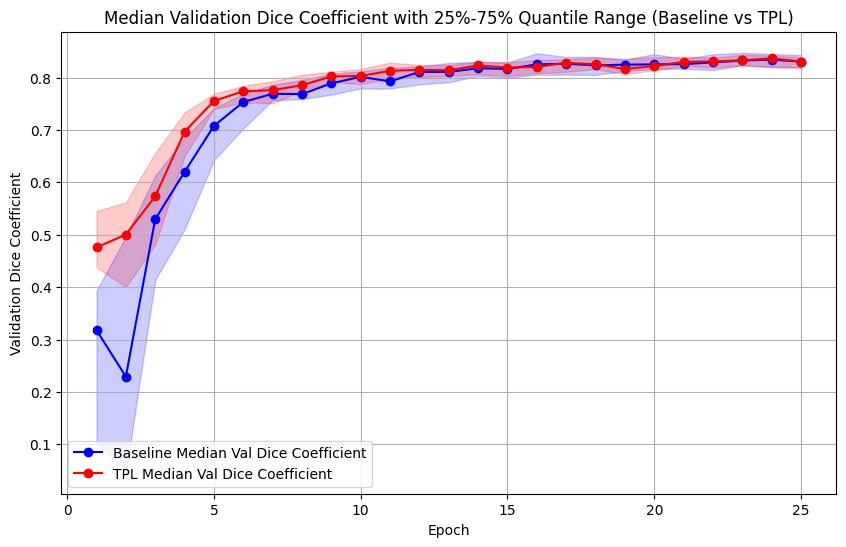}
        \subcaption{Dice score ($128 \times 128$, 64 channels)}
    \end{minipage}
    \hfill
    \begin{minipage}{0.48\textwidth}
        \centering
        \includegraphics[width=\textwidth]{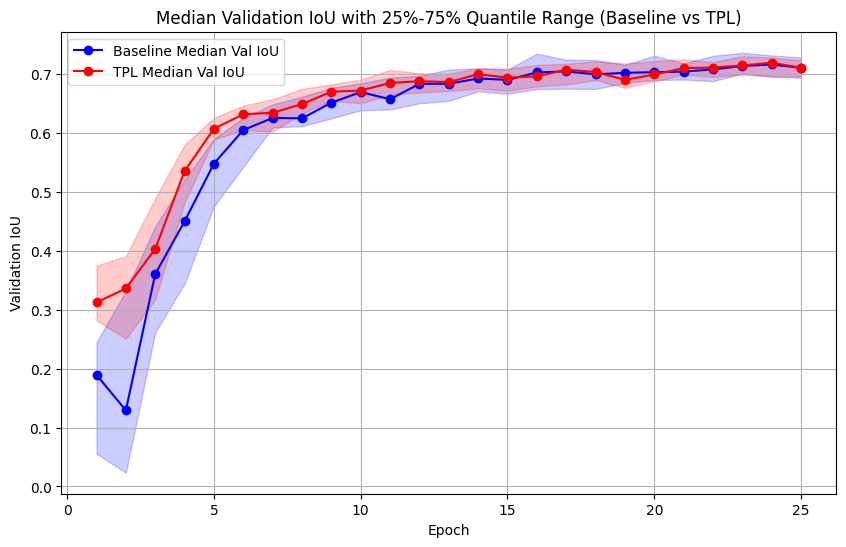}
        \subcaption{IoU score ($128 \times 128$, 64 channels)}
    \end{minipage}
    \caption{
    Validation Dice coefficient and IoU per epoch for input size $128 \times 128$. 
The values at each epoch represent the median from 30 repeated experiments, with the shaded region indicating the range between the 25th and 75th percentiles (quantiles). 
The top row corresponds to cases where the number of channels in the first convolutional layer is 16, while the bottom row corresponds to cases with 64 channels.}
    \label{fig:kvasir_128}
\end{figure}

Figure~\ref{fig:kvasir_128} presents 4 plots displaying the Dice coefficients and IoU values for the validation data across 25 epochs. 
The top row corresponds to the model with 16 channels in the first convolutional layer, while the bottom row corresponds to the model with 64 channels. 
Each value represents the median result from 30 repeated experiments, with the shaded region indicating the range between the 25th and 75th percentiles (quantiles).
Below are our observations:

\begin{itemize}
    \item 
    For the model with 16 channels (top row), the TPL model slightly outperforms the baseline model in terms of Dice coefficients and IoU scores during the first few epochs. 
    However, as training progresses, the performance gap between the two models narrows.
    \item 
    For the model with 64 channels (bottom row), we observe a similar trend with the performance gap narrowing in later stages. 
    However, during the early epochs, the TPL model shows faster improvement compared to the baseline, with a narrower quantile range, suggesting that the TensorProjection layer contributes to greater stability in learning for this model. 
\end{itemize}

We then conducted an additional experiment to further test the effectiveness of the TensorProjection layer for dimension reduction. 
In this experiment, the baseline model remained unchanged, but we introduced a slight adjustment to the TPL model.
In the previous experiment, the TensorProjection layer compressed the input features from $16\times 16$ (with 128 or 512 channels) to $8\times 8$. 
However, in this experiment, we compressed the input features further, reducing their size from $16\times 16$ to a smaller dimension of $4\times 4$, and then upsampled them back to $16\times 16$ using the TransposeTensorProjection layer.

\begin{figure}[h!]
    \centering
    \begin{minipage}{0.48\textwidth}
        \centering
        \includegraphics[width=\textwidth]{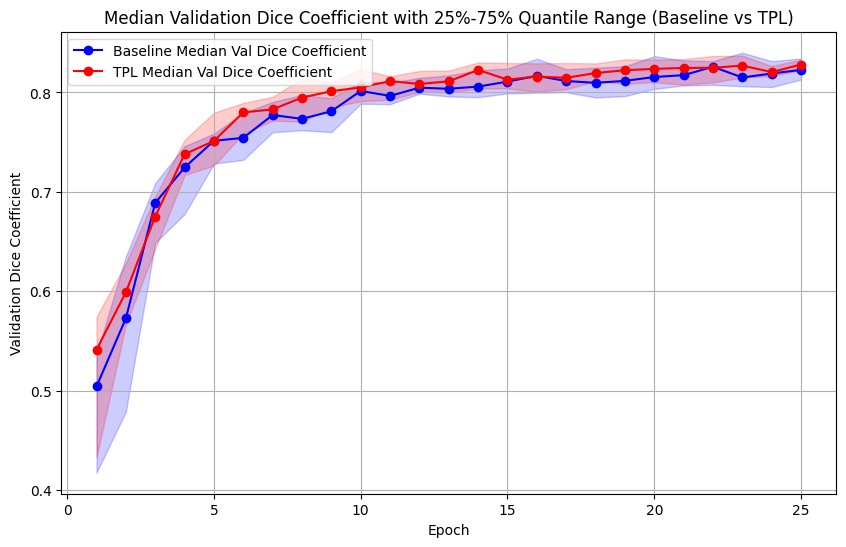}
        \subcaption{Dice score ($128 \times 128$, 16 channels) more aggressive compression in TPL model}
    \end{minipage}
    \hfill
    \begin{minipage}{0.48\textwidth}
        \centering
        \includegraphics[width=\textwidth]{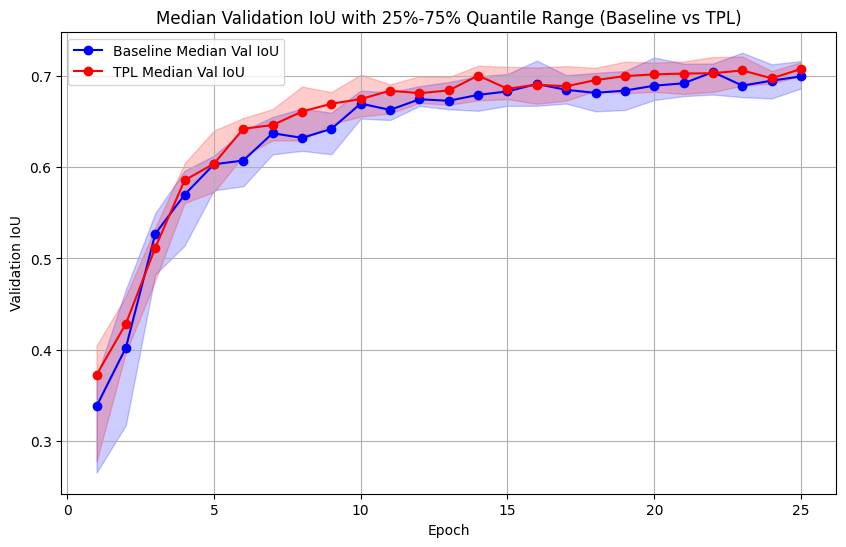}
        \subcaption{IoU score ($128 \times 128$, 16 channels) more aggressive compression in TPL model}
    \end{minipage}

    \vspace{0.15cm}

    \begin{minipage}{0.48\textwidth}
        \centering
        \includegraphics[width=\textwidth]{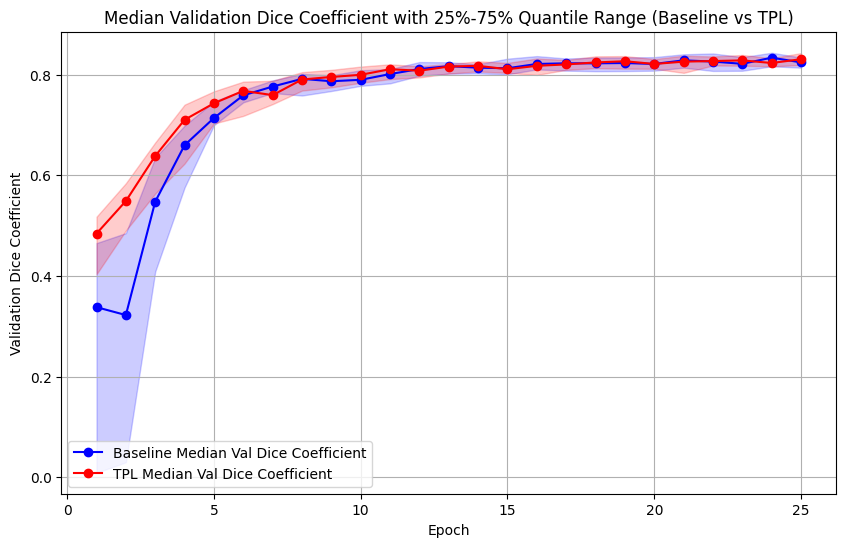}
        \subcaption{Dice score ($128 \times 128$, 64 channels) more aggressive compression in TPL model}
    \end{minipage}
    \hfill
    \begin{minipage}{0.48\textwidth}
        \centering
        \includegraphics[width=\textwidth]{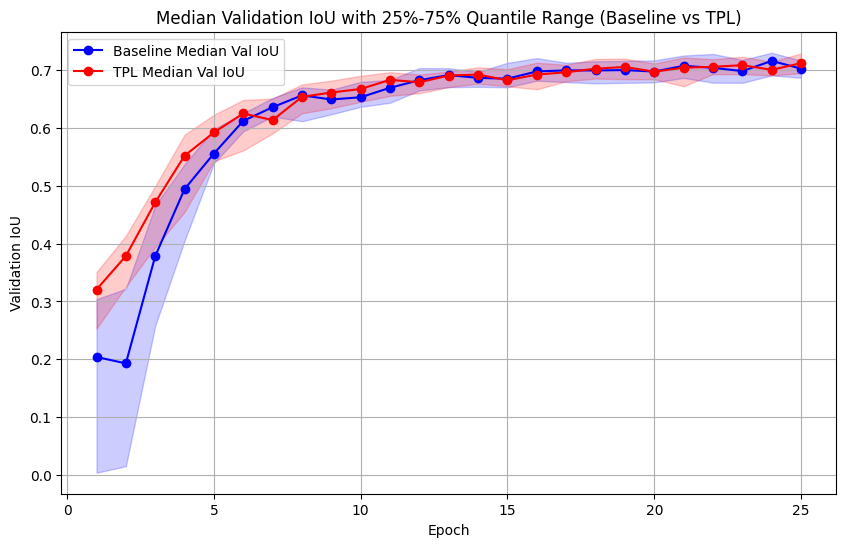}
        \subcaption{IoU score ($128 \times 128$, 64 channels) more aggressive compression in TPL model}
    \end{minipage}
    \caption{
    Validation Dice coefficient and IoU per epoch for input size $128 \times 128$ in the case where the output size of the TensorProjection layer is reduced further. 
    The values at each epoch represent the median from 30 repeated experiments, with the shaded region indicating the range between the 25th and 75th percentiles (quantiles).
    The top row corresponds to models with 16 channels in the first convolutional layer, and the bottom row corresponds to models with 64 channels. 
    In contrast to previous experiments, in this case, the TensorProjection layer compresses the feature maps from $16 \times 16$ down to $4 \times 4$ before being upsampled back to $16 \times 16$ using the Transpose TensorProjection layer. 
    This setup tests the impact of a more aggressive dimension reduction on model performance.
    }
    \label{fig:kvasir_128_unfair}
\end{figure}

The results of this experiment are summarized in Figure~\ref{fig:kvasir_128_unfair}. 
Although no changes were made to the baseline model, the values are based on newly conducted experiments, so some slight variations from previous results may be observed.
Our observations based on this figure are as follows:

\begin{itemize}
\item 
The TPL model demonstrates faster performance improvement than the baseline during the initial epochs, particularly notable with 64 channels. 
\item 
While both models ultimately converge to similar performance levels, it is noteworthy that the TPL model maintained overall performance, even when compressing to smaller dimensions in the decoder. 
This finding supports the effectiveness of dimensiona reduction using the TensorProjection layer.
\end{itemize}

We also conducted experiments with input sizes of $160\times160$ and $192\times192$. 
The results of these experiments are presented in Appendix \ref{appendix:additional_results}.

\clearpage


%
%
%
%

\section{Conclusion}\label{sec:Conclusion}

In this study, we proposed a dimension reduction method tailored for tensor-structured data within deep neural networks.  
In Section~\ref{sec:TensorProjectionLayer}, we derived the gradients of the loss function with respect to the parameters associated with the TensorProjection layer.  
In Section~\ref{sec:examples}, we incorporated the TensorProjection layer into models for image classification and segmentation, and evaluated its effectiveness by comparing it with baseline models.  
For the image segmentation task, we also introduced a TransposeTensorProjection layer in the decoder part of the U-Net models.  
Below, we summarize the key features of the TensorProjection layer:
\begin{itemize}
    \item \textbf{Dimension reduction effect and representational capacity}:
The TensorProjection layer outperformed traditional dimension reduction methods (such as spatial summarization through pooling or channel reduction using convolution) in our experiments.  
In particular, incorporating the TensorProjection layer resulted in faster model training in Sections \ref{sec:retinal_oct} and \ref{sec:chest_x_ray}.  
(A similar effect was also observed during the early epochs in the experiment in Section \ref{sec:kvasir_segmentation}.)  
This improvement is likely because the TensorProjection layer derives projection directions based on a supervised learning criterion, enhancing its representational capacity compared to pooling layers in appropriate contexts.  
Pooling layers effectively summarize data by leveraging the similarities among adjacent pixel values in the early layers, but in later layers, they may overcompress information.  
In such cases, the TensorProjection layer is expected to reduce dimensions while facilitating more efficient information transmission.  
It is also noteworthy that in Section \ref{sec:kvasir_segmentation}, when the number of channels was increased, the baseline model exhibited performance variability during the early epochs, whereas the TPL model maintained relatively stable performance.

   \item \textbf{Flexible dimension adjustment while preserving the tensor structure}: \\
A major feature of the TensorProjection layer is its ability to flexibly adjust the dimensions of data along each axis while maintaining the tensor structure. 
In MaxPooling layers, the output dimensions are fixed based on the stride and pool size, and can only summarize data in the spatial dimensions (height and width). 
Similarly, Global Average Pooling layers can only reduce dimensions along the channel axis. 
However, the TensorProjection layer allows for dimension reduction across all axes, including both spatial and channel dimensions, providing practitioners with a more flexible approach to model design. 
This flexibility enables the TensorProjection layer to offer more adaptable and versatile dimension reduction compared to traditional methods.

    \item \textbf{Overfitting risk and its control}:\\  
    The TensorProjection layer incorporates trainable model parameters, which introduces a potential risk of overfitting.
In our experiments, where we compared the TensorProjection layer and pooling layers with identical output dimensions, we observed that while the TensorProjection layer resulted in faster training, it also showed earlier signs of overfitting.
This can be attributed to the fact that both layers were set with equal output dimensions.
That said, the TensorProjection layer has the potential to mitigate overfitting by further reducing output dimensions.
By decreasing dimension, the number of model parameters is reduced, which in turn limits model complexity and helps prevent overfitting.
Thus, reducing dimensions appropriately allows the TensorProjection layer to maintain high representational power while effectively preventing overfitting.
 \end{itemize}

%
%

\section*{Data Availability}

The datasets used for the examples in Section \ref{sec:examples} are publicly available for download from Kaggle. 
For details on the specific datasets and their download links, please refer to the respective subsections in Section \ref{sec:examples}. 
Each subsection includes the corresponding URL and citations as specified on the Kaggle pages.

Additionally, the TensorProjection layer code that we implemented will be made publicly available on GitHub. 
The code will be accessible at \href{https://github.com/senyuan-juncheng/TensorProjectionLayer}{https://github.com/senyuan-juncheng/TensorProjectionLayer}.

\section*{Author Contributions}

\begin{itemize}\itemsep=4pt
\item \textbf{Toshinari Morimoto}: Formal analysis, Data curation, Methodology, Software, Validation, Visualization, Writing – original draft.  
\item \textbf{Su-Yun Huang}: Conceptualization, Funding acquisition, Methodology, Supervision, Validation, Writing – review and editing.
\end{itemize}

\section*{Statements and Declarations}
\begin{itemize}\itemsep=4pt
\item \textbf{Competing Interests}: 
The authors declare that they have no competing interests.
\item \textbf{Use of Generative AI and AI-assisted technologies in the writing process}: 
During the preparation of this work, the authors utilized ChatGPT to ensure grammatical accuracy, translate content, and improve the natural flow of the English text. 
After using this service, the authors carefully reviewed and revised the content as necessary and take full responsibility for the final version of the published article.
\end{itemize}

\section*{Acknowledgement}
This research was partially supported by the National Science and Technology Council, Taiwan, under grant no. MOST 110-2118-M-001-007-MY3.

%
%

\bibliography{references}


%
%

\begin{appendices}
%
%

\section{Proof for Proposition \ref{prop:gradients_of_TensorProjectionLayer}}\label{appendix:proof}

\subsection{Equations (\ref{eq:dLdvecXi}) and (\ref{eq:dLdvecWk})}
The first equality in Equation (\ref{eq:dLdvecXi}) is obtained by applying the Chain Rule (see \ref{appendix:chainrule} for more details). 
The second equality in Equation (\ref{eq:dLdvecXi}) follows from the vectorization operator, $\vect(\cdot)$, being applied to both sides of Equation (\ref{eq:forward_propagation}).
It is important to note the well-known result:
\[
\vect(X \times_1 A_1 \times_2 A_2 \cdots \times_k A_k) = \left(A_k \otimes \cdots \otimes A_1\right) \vect(X).
\]
Equation (\ref{eq:dLdvecWk}) also follows directly from the Chain Rule. 
In this manner, we decompose $\frac{\partial L}{\partial \vect(W_k)^\top}$ into a product of three gradients. 
Given that $\frac{\partial L}{\partial \vect(\Z_i)^\top}$ is known, we now focus on how to compute $ \frac{\partial \vect(\Z_i)}{\partial \vect(U_k)^\top}$ and $\frac{\partial \vect(U_k)}{\partial \vect(W_k)^\top}$ in the subsequent sections.

\subsection{Equations (\ref{eq:dvecZidvecU1}), (\ref{eq:dvecZidvecU2}) and (\ref{eq:dvecZidvecU3})}

The derivations of (\ref{eq:dvecZidvecU1}), (\ref{eq:dvecZidvecU2}), and (\ref{eq:dvecZidvecU3}) follow a similar process. 
Therefore, it suffices to discuss the derivation of (\ref{eq:dvecZidvecU1}).
By unfolding $\Z_i$ along the first mode (axis or dimension), we have
\[
\Z_{i_{(1)}} = U_1^\top \X_{i_{(1)}} \left(U_3 \otimes U_2\right),
\]
where $\Z_{i_{(1)}}$ and $\X_{i_{(1)}}$ denote the tensor unfoldings of $\Z_i$ and $\X_i$ along the first mode, respectively. 
Note that there are different conventions for mode-$k$ tensor unfolding, and in this paper, we follow the definition by \citet{Kolda2009}.
Since $\Z_{i_{(1)}} = U_1^\top \X_{i_{(1)}} (U_3 \otimes U_2) = I_{q_1} U_1^\top \X_{i_{(1)}} (U_3 \otimes U_2)$, we can derive:
\begin{eqnarray*}
\vect\left(\Z_{i_{(1)}}\right) &=& \vect\left(I_{q_1} U_1^\top \X_{i_{(1)}} \left(U_3 \otimes U_2\right)\right) 
=  \left[ \left( \left( U_3^\top \otimes U_2^\top \right) \X_{i_{(1)}}^\top \right) \otimes I_{q_1} \right] \vect(U_1^\top) \\ 
&=&  \left[ \left( \left( U_3^\top \otimes U_2^\top \right) \X_{i_{(1)}}^\top \right) \otimes I_{q_1} \right] K_{p_1, q_1} \vect(U_1),
\end{eqnarray*}
where $K_{p_1, q_1}$ is the commutation matrix. This follows from the known identity:
\[
\vect(AXB) = (B^\top \otimes A) \vect(X).
\]
A commutation matrix $K_{m,n}$ satisfies $\vect(A^\top) = K_{m,n} \vect(A)$ for any $m \times n$ matrix $A$. 
For more details on commutation matrices, refer to \cite{Xu2018}.
From the result above, we obtain:
\[
\frac{\partial \vect(\Z_{i_{(1)}})}{\partial \vect(U_1)^\top} = \left[ \left( \left( U_3^\top \otimes U_2^\top \right) \X_{i_{(1)}}^\top \right) \otimes I_{q_1} \right] K_{p_1, q_1}.
\]
Additionally, there exists a permutation matrix $P^{(1)}_{q_1,q_2,q_3}$ such that:
\[
\vect(\Z_i) = P^{(1)}_{q_1,q_2,q_3} \vect(\Z_{i_{(1)}}),
\]
which depends only on $q_1, q_2, q_3$. 
By applying the chain rule, we then have:
\begin{eqnarray*}
\frac{\partial \vect(\Z_i)}{\partial \vect(U_1)^\top} &=& \frac{\partial \vect(\Z_i)}{\partial \vect(\Z_{i_{(1)}})^\top} \frac{\partial \vect(\Z_{i_{(1)}})}{\partial \vect(U_1)^\top} \\ 
&=& P^{(1)}_{q_1,q_2,q_3} \left[ \left( \left( U_3^\top \otimes U_2^\top \right) \X_{i_{(1)}}^\top \right) \otimes I_{q_1} \right] K_{p_1, q_1}.
\end{eqnarray*}
Thus, we obtain Equation (\ref{eq:dvecZidvecU1}). 
The derivations of Equations (\ref{eq:dvecZidvecU2}) and (\ref{eq:dvecZidvecU3}) can be proven in a similar manner.

\subsection{Equation (\ref{eq:vecUkvecWk})}

Next, we verify Equation (\ref{eq:vecUkvecWk}). 
Note that $U_k = W_k G_k$. 
Now, consider a small variation where $W_k \to W_k + \Delta W_k$. Since both $G_k$ and $U_k$ are functions of $W_k$, we have:
\[
U_k + \Delta U_k = (W_k + \Delta W_k)(G_k + \Delta G_k).
\]
By neglecting the second-order term $\Delta W_k \Delta G_k$, we obtain:
\[
\Delta U_k = \Delta W_k G_k + W_k \Delta G_k = I_{p_k} \Delta W_k G_k + W_k \Delta G_k I_{q_k}.
\]
Applying the vectorization operator to both sides gives:
\begin{eqnarray*}
\vect(\Delta U_k) &=& \vect(I_{p_k} \Delta W_k G_k) + \vect(W_k \Delta G_k I_{q_k}) \\ 
&=& \left(G_k^\top \otimes I_{p_k}\right) \vect(\Delta W_k) + \left(I_{q_k} \otimes W_k\right) \vect(\Delta G_k).
\end{eqnarray*}
Since the operators $\Delta$ and $\vect$ can be interchanged, we have:
\[
\Delta \vect(U_k) = \left(G_k^\top \otimes I_{p_k}\right) \Delta \vect(W_k) + \left(I_{q_k} \otimes W_k\right) \Delta \vect(G_k).
\]
This directly leads to the desired result in Equation (\ref{eq:vecUkvecWk}).

\subsection{Equations (\ref{eq:vecGkvecWk}), (\ref{eq:vecGkvecMk}) and (\ref{eq:vecMkvecWk})}

Equation (\ref{eq:vecUkvecWk}) still contains an unsolved part, namely Equation (\ref{eq:vecGkvecWk}). 
By the Chain Rule, we can decompose Equation (\ref{eq:vecGkvecWk}) into two parts: Equations (\ref{eq:vecGkvecMk}) and (\ref{eq:vecMkvecWk}).

We first address Equation (\ref{eq:vecGkvecMk}). 
For simplicity, let $M$ be a $q \times q$ symmetric positive definite matrix, and define $G \eqdef M^{-1/2}$. 
We aim to compute $\frac{\partial \vect(G)}{\partial \vect(M)^\top}$. 
Let $\Delta M$ be a small variation of $M$, where $\Delta M$ is also symmetric and positive definite.
We then have:
\begin{equation}
\Delta G = \lim_{\epsilon \rightarrow 0} \frac{(M+\epsilon \Delta M)^{-1/2} - M^{-1/2}}{\epsilon} \label{eq:dG}.
\end{equation}
By simple manipulation, we can express:
\begin{eqnarray}
(M+\epsilon \Delta M)^{-1/2} &=& \Bigl \{ M^{1/2} \left (I_q + \epsilon M^{-1/2} \Delta M M^{-1/2} \right ) M^{1/2} \Bigr \}^{-1/2} \\
\label{eq:M_add_epsilonDeltaM_invsqrt_note1} &=& M^{-1/4} \left(I_q + \epsilon M^{-1/2} \Delta M M^{-1/2}\right)^{-1/2} M^{-1/4} \\
\label{eq:M_add_epsilonDeltaM_invsqrt_note2} &=& M^{-1/4} \left(I_q + \epsilon V \Lambda V^\top \right)^{-1/2} M^{-1/4}   \\
\label{eq:M_add_epsilonDeltaM_invsqrt_note3} &=& M^{-1/4} \left(V (I_q + \epsilon \Lambda) V^\top \right)^{-1/2} M^{-1/4}  \\
\label{eq:M_add_epsilonDeltaM_invsqrt_note4} &=& M^{-1/4} V \left (I_q + \epsilon \Lambda \right)^{-1/2} V^\top  M^{-1/4} .
\end{eqnarray}
Additionally, we note the following important points:
\begin{itemize}
  \item (\ref{eq:M_add_epsilonDeltaM_invsqrt_note1}) to (\ref{eq:M_add_epsilonDeltaM_invsqrt_note2}):  We consider eigenvalue decomposition of the symmetric matrix: \[M^{-1/2} \Delta M M^{-1/2} = V \Lambda V^\top.\]
  \item (\ref{eq:M_add_epsilonDeltaM_invsqrt_note2}) to (\ref{eq:M_add_epsilonDeltaM_invsqrt_note3}):  Expanding the expression inside $(\ldots)^{-1/2}$ in  (\ref{eq:M_add_epsilonDeltaM_invsqrt_note3}) results in (\ref{eq:M_add_epsilonDeltaM_invsqrt_note2}).
  (Here note that $V V^\top = I_q$ because $V$ is an orthogonal matrix.)
  \item (\ref{eq:M_add_epsilonDeltaM_invsqrt_note3}) to (\ref{eq:M_add_epsilonDeltaM_invsqrt_note4}): The following equation can be easily verified:
  \[(V \left (I_q + \epsilon \Lambda \right)^{-1/2} V^\top)^2 = V \left (I_q + \epsilon \Lambda \right)^{-1}V^\top = (V \left (I_q + \epsilon \Lambda \right)V^\top)^{-1}.\]
  Thus, by taking the square root of the matrices on both the left-hand side and the right-hand side, the right-hand side becomes (\ref{eq:M_add_epsilonDeltaM_invsqrt_note3}), while the left-hand side corresponds to (\ref{eq:M_add_epsilonDeltaM_invsqrt_note4}).
\end{itemize}
Also note that
\[
M^{-1/2} = M^{-1/4} V I_q V^\top M^{-1/4}.
\]
By using these relations, we can simplify (\ref{eq:dG}) as:
\[
M^{-1/4} V \left\{ \lim_{\epsilon \rightarrow 0} \frac{ \left(I_q + \epsilon \Lambda \right)^{-1/2} - I_q }{\epsilon} \right \} V^\top  M^{-1/4}.
\]
It is straightforward to verify that:
\[
\lim_{\epsilon \rightarrow 0} \frac{ \left(I_q + \epsilon \Lambda \right)^{-1/2} - I_q }{\epsilon} = -\frac{1}{2} \Lambda,
\]
since each element on the diagonal can be computed as:
\[
\lim_{\epsilon \rightarrow 0}\frac{1}{\epsilon} \left(\frac{1}{\sqrt{1+\epsilon \lambda_i}} - 1\right) = -\frac{1}{2} \lambda_i.
\]
As a result, we obtain:
\begin{eqnarray*}
\Delta G &=& -\frac{1}{2} M^{-1/4} V \Lambda V^\top  M^{-1/4} 
  = -\frac{1}{2} M^{-1/4} M^{-1/2} \Delta M M^{-1/2}  M^{-1/4} 
  = -\frac{1}{2} M^{-3/4} \Delta M M^{-3/4} .
\end{eqnarray*}
By vectoring the both sides, we have
\[
\vect (\Delta G) = -\frac{1}{2} \left (M^{-3/4} \otimes M^{-3/4}\right) \vect(\Delta M),
\]
from which we immediately obtain Equation  (\ref{eq:vecGkvecMk}).

Finally we verify  Equation (\ref{eq:vecMkvecWk}).
Similarly, consider a small variation $W_k \to W_k + \Delta W_k$.
We have:
\[
M_k + \Delta M_k = (W_k + \Delta W_k)^\top(W_k + \Delta W_k).
\]
By ignoring the second-order term $\Delta W_k^\top \Delta W_k$, we obtain:
\[
\Delta M_k =  W_k^\top \Delta W_k + (\Delta W_k)^\top W_k = W_k^\top \Delta W_k I_{q_k} + I_{q_k} (\Delta W_k)^\top W_k.
\]
Vectorizing both sides gives:
\begin{eqnarray*}
\vect (\Delta M_k) &=& \vect\left(W_k^\top \Delta W_k I_{q_k}\right) + \vect\left(I_{q_k} (\Delta W_k)^\top W_k\right) \\
&=& \left(I_{q_k} \otimes W_k^\top\right) \vect\left(\Delta W_k\right) + \left(W_k^\top \otimes I_{q_k}\right) \vect\left(\left(\Delta W_k\right)^\top\right) \\
&=& \left(I_{q_k} \otimes W_k^\top\right) \vect\left(\Delta W_k\right) + \left(W_k^\top \otimes I_{q_k}\right) K_{p_k, q_k} \vect\left(\Delta W_k\right) .
\end{eqnarray*}
Now the proof is complete.

%
%

\section{Gradient}\label{appendix:gradient}

This section reviews basic mathematical concepts such as the gradient and chain rule. 
While these are fundamental, we include them here to clarify our notation and avoid any potential confusion.

\subsection{Gradient for a scalar function}

Let $\bm{x} \in \mathbb{R}^p$, and consider a scalar function $f \eqdef f(\bm{x}): \mathbb{R}^p \to \mathbb{R}$, where $\bm{x} = \left(x_1, x_2, \ldots, x_p \right)^\top$.
In this paper, we define the gradient of $f$ with respect to \( \bm{x} \) as:
\[
\frac{\partial f}{\partial \bm{x}^\top} \eqdef \left(\frac{\partial f}{\partial x_1}, \frac{\partial f}{\partial x_2}, \ldots, \frac{\partial f}{\partial x_p}\right).
\]
Here, the gradient is represented as a row vector. 
This is convenient when applying the chain rule, as it allows for a more straightforward combination of gradients when differentiating composite functions.

\subsection{Gradient for a vector-valued function}

Let $\bm{g} \eqdef \left(g_1(\bm{x}), g_2(\bm{x}), \ldots, g_d(\bm{x}) \right)^\top : \mathbb{R}^p \to \mathbb{R}^d$ be a vector-valued function, where $\bm{x} \eqdef (x_1, x_2, \ldots, x_p)^\top \in \mathbb{R}^p$.
In this context, the derivative of $\bm{g}$ with respect to $\bm{x}$ is expressed as a matrix (Jacobian matrix), where each row corresponds to the gradient of one component of  $\bm{g}$. 
We define it as follows:
\[
\frac{\partial \bm{g}}{\partial \bm{x}^\top} \eqdef
\begin{pmatrix} 
\frac{\partial g_1(\bm{x})}{\partial x_1} & \frac{\partial g_1(\bm{x})}{\partial x_2} & \cdots & \frac{\partial g_1(\bm{x})}{\partial x_p}  \\ 
\frac{\partial g_2(\bm{x})}{\partial x_1} & \frac{\partial g_2(\bm{x})}{\partial x_2} & \cdots & \frac{\partial g_2(\bm{x})}{\partial x_p}  \\
\vdots & \vdots & \ddots & \vdots  \\ 
\frac{\partial g_d(\bm{x})}{\partial x_1} & \frac{\partial g_d(\bm{x})}{\partial x_2} & \cdots & \frac{\partial g_d(\bm{x})}{\partial x_p}  
\end{pmatrix}.
\]

\subsection{Chain rule}\label{appendix:chainrule}

Let $\bm{h} \eqdef \bm{h}(\bm{u})$ be a vector-valued (or a scalar) function with respect to a vector $\bm{u} = (u_1,\ldots,u_r)^\top$.
Suppose that $\bm{u} \eqdef \bm{u}(\bm{x})$ is a function of another vector $\bm{x} = (x_1, \ldots, x_p)^\top$.
Then, the following equation holds:
\[
\frac{\partial\bm{h}}{\partial \bm{x}^\top} = \frac{\partial \bm{h}}{\partial \bm{u}^\top} \frac{\partial \bm{u}}{\partial\bm{x}^\top}.
\]

As a simple example, let $h(\bm{u}) \eqdef u_1 u_2$ be a scalar function with respect to $\bm{u}$, and let $u_1 \eqdef (x_1 + x_2), u_2 \eqdef (x_2 + x_3)$ be functions with respect to $\bm{x}=(x_1,x_2,x_3)^\top$.
By applying the Chain Rule as shown above, we have:
\[
\frac{\partial h}{\partial \bm{u}^\top} \frac{\partial \bm{u}}{\partial \bm{x}^\top} = \left(u_2 , u_1\right) \begin{pmatrix} 1 & 1 & 0 \\ 0 & 1 & 1 \end{pmatrix}  = (u_2, u_1 + u_2, u_1) = (x_2 + x_3, x_1 + 2x_2 + x_3, x_1 + x_2).
\]

Alternatively, since $h \stackrel{\rm eq}{=} (x_1 + x_2)(x_2 + x_3)$, we can directly obtain
\[
\frac{\partial h}{\partial \bm{x}^\top} = \left(x_2 + x_3, x_1 + 2x_2 + x_3, x_1 + x_2\right).
\]

%
%

\section{Vectorization operator}\label{appendix:vec_operation}

A vectorization operator transforms a tensor (or a matrix) into a vector.
For example, suppose that \( A \) is a matrix of size \( m \times n \).
Then
\[ \vect(A) \eqdef \left(a_{11}, a_{21}, \ldots, a_{m1}, a_{12}, a_{22}, \ldots, a_{m2}, \ldots, a_{1n} , a_{2n} , \ldots, a_{mn} \right)^\top. \]
Similarly, if \( \mathcal{A} \) is a tensor of size \( p \times q \times r \), then
\[\vect(\mathcal{A}) \eqdef \left( a_{111}, a_{211}, \ldots, a_{p11}, a_{121}, a_{221}, \ldots, a_{p21}, \ldots, a_{pq1}, a_{112}, a_{212}, \ldots, a_{p12}, \ldots, a_{pqr} \right)^\top.\]

In Python (NumPy), reshaping into \( pqr \times 1 \) with the option \texttt{order='F'} is equivalent to the vectorization operation. 
By default, NumPy arranges the elements in \texttt{'C'} order, where the last axis is traversed first, but using \texttt{order='F'} traverses the first axis first, which matches the behavior of the vectorization operator.

For more details on the properties of the vec operator, please refer to \cite{Petersen2012}.

%
%

\section{Mode-$k$ product}\label{appendix:k-mode-product}

\subsection{Definition of  mode-$k$ product}

We provide the definition of the mode-$k$ product between a third-order tensor and a matrix. 
Let $\X$ be a tensor of size $p \times q \times r$, and let $A$, $B$, and $C$ be matrices of size $p' \times p$, $q' \times q$, and $r' \times r$, respectively. 
The mode-$k$ products $\mathcal{T}_1 \eqdef \X \times_1 A$, $\mathcal{T}_2 \eqdef \X \times_2 B$, and $\mathcal{T}_3 \eqdef \X \times_3 C$ are defined as follows. 
The resulting tensors $\{\mathcal{T}_k\}_{k=1}^3$ have dimensions $p' \times q \times r$, $p \times q' \times r$, and $p \times q \times r'$, respectively:
\begin{eqnarray*}
(\mathcal{T}_1)_{i,j,k} &=& \sum_{\ell=1}^p A_{i,\ell} \X_{\ell, j,k}, \\
(\mathcal{T}_2)_{i,j,k} &=& \sum_{\ell=1}^q B_{j,\ell} \X_{i, \ell,k}, \\
(\mathcal{T}_3)_{i,j,k} &=& \sum_{\ell=1}^r C_{k,\ell} \X_{i,j,\ell}.
\end{eqnarray*}

\subsection{Alternative definition of mode-$k$ product}

The mode-$k$ product can also be defined using the concepts of unfolding and folding. 
While there are several different definitions of unfolding and folding, the definition of the mode-$k$ product remains the same regardless of the specific approach. 
We follow the definitions of unfolding and folding as described by \citet{Kolda2009}, and the URL in the footnote also provides a helpful reference.\footnote{\url{https://jeankossaifi.com/blog/unfolding.html}}

\subsubsection{Unfolding and folding}\label{appendix:unfolding_folding}

Let $\X$ be a tensor of size $p_1 \times p_2 \times p_3$. 
Unfolding is the operation of converting a tensor into a matrix by first permuting its axes and then reshaping it. 
The unfolding of $\X$ along the $k$-th mode, where $k=1,2,3$, is denoted by $\X_{(k)}$. 
In the reshape operation, we assume that the index of the first mode (axis) changes faster than the third (or last) mode, consistent with the vectorization method defined earlier. 
In NumPy's \texttt{reshape} function, the option \texttt{order='F'} needs to be used to match this definition.
The definitions of $\X_{(k)}$ for each mode are specifically given as follows:
\begin{align*}
  \X_{(1)} &\eqdef \mathrm{reshape}(\mathrm{permute}(\X, [1,2,3]),[p_1, p_2p_3]) ,\\
  \X_{(2)} &\eqdef \mathrm{reshape}(\mathrm{permute}(\X, [2,1,3]),[p_2, p_3p_1]), \\
  \X_{(3)} &\eqdef \mathrm{reshape}(\mathrm{permute}(\X, [3,1,2]),[p_3, p_1p_2]).
\end{align*}

Folding is the operation that reshapes and permutes a matrix into a tensor form based on specified dimensions. 
In the following equations, $p_1, p_2, p_3$ represent the target dimensions of the tensor. 
As a result, folding can be viewed as the inverse of the unfolding process, but its primary purpose is simply to transform the matrix into the desired tensor shape.
\begin{align*}
    \text{if } k = 1: & \quad \X = \mathrm{permute}(\mathrm{reshape}(\X_{(k)}, [p_1, p_2, p_3]), [1, 2, 3]), \\
    \text{if } k = 2: & \quad \X = \mathrm{permute}(\mathrm{reshape}(\X_{(k)}, [p_2, p_1, p_3]), [2, 1, 3]), \\
    \text{if } k = 3: & \quad \X = \mathrm{permute}(\mathrm{reshape}(\X_{(k)}, [p_3, p_1, p_2]), [2, 3, 1]).
\end{align*}

\subsubsection{Mode-$k$ product based on folding and unfolding}

Let $\X$ be a tensor of size $p_1 \times p_2 \times p_3$, and let $M_k$ be a matrix of size $q_k \times p_k$. 
The mode-$k$ product, denoted by $\X \times_k M_k$, results in a tensor of size $q_1 \times p_2 \times p_3$ (if $k=1$), $p_1 \times q_2 \times p_3$ (if $k=2$), or $p_1 \times p_2 \times q_3$ (if $k=3$).
The mode-$k$ product is also computed in the following steps:
\begin{description}
  \item[Step 1.] Compute the unfolding $\X_{(k)}$ along the $k$-th mode.
  \item[Step 2.] Multiply $M_k$ by the matrix $\X_{(k)}$. 
  \item[Step 3.] Fold the resulting matrix $M_k \X_{(k)}$ back into a tensor along the $k$-th mode. 
  The resulting tensor will have the shape $q_1 \times p_2 \times p_3$, $p_1 \times q_2 \times p_3$, or $p_1 \times p_2 \times q_3$, depending on the value of $k$.
\end{description}

%
%
\section{Additional experimental results}\label{appendix:additional_results}

\begin{figure}[h!]
    \centering
    \begin{minipage}{0.48\textwidth}
        \centering
        \includegraphics[width=\textwidth]{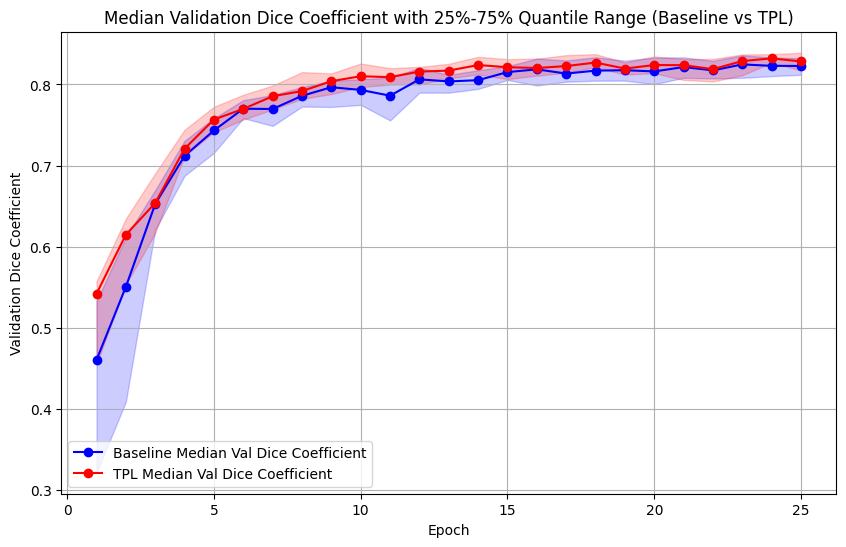}
        \subcaption{Dice score ($160 \times 160$, 16 channels)}
    \end{minipage}
    \hfill
    \begin{minipage}{0.48\textwidth}
        \centering
        \includegraphics[width=\textwidth]{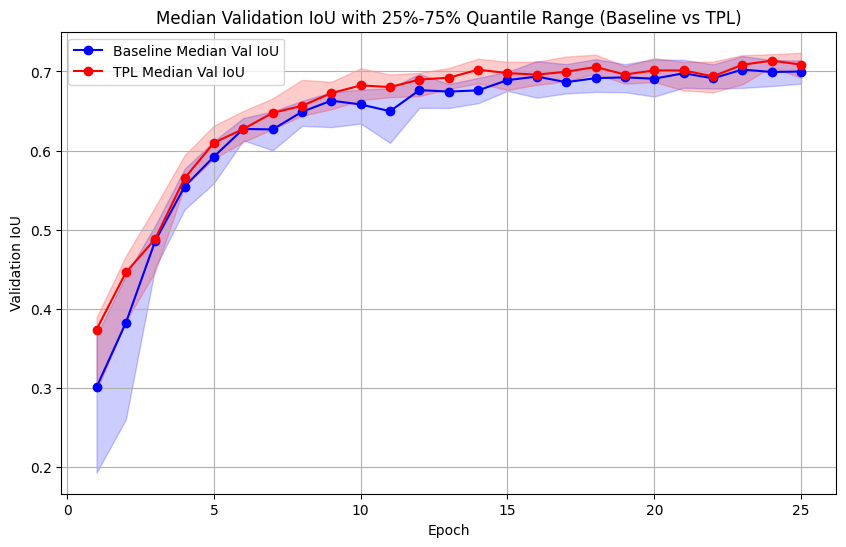}
        \subcaption{IoU score ($160 \times 160$, 16 channels)}
    \end{minipage}

    \vspace{0.3cm} 

    \begin{minipage}{0.48\textwidth}
        \centering
        \includegraphics[width=\textwidth]{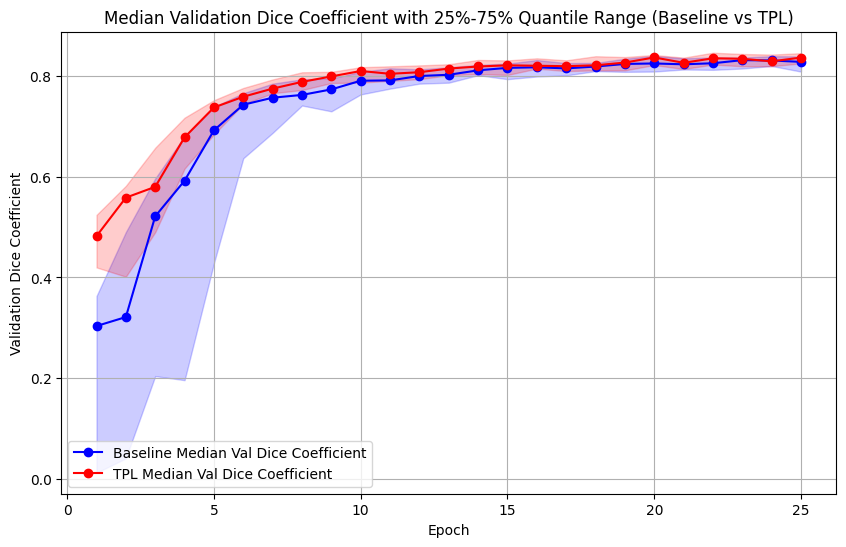}
        \subcaption{Dice score $160 \times 160$, 64 channels)}
    \end{minipage}
    \hfill
    \begin{minipage}{0.48\textwidth}
        \centering
        \includegraphics[width=\textwidth]{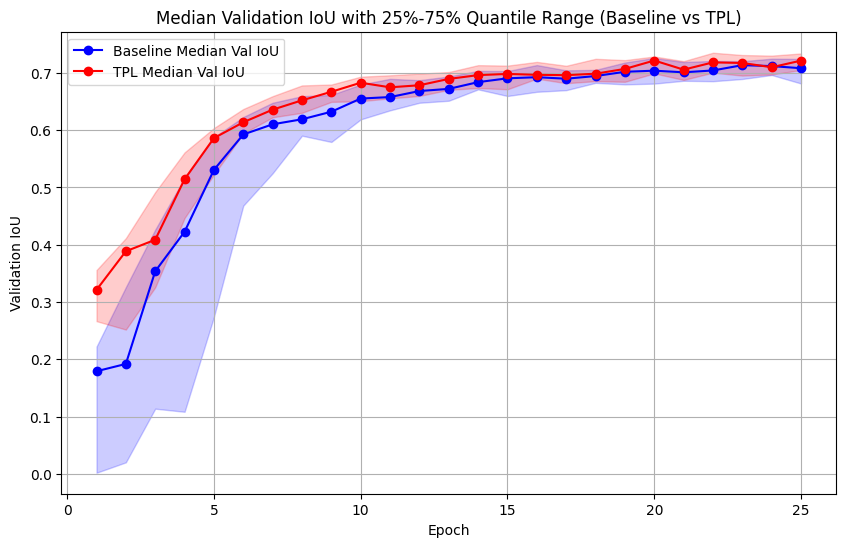}
        \subcaption{IoU score $160 \times 160$, 64 channels)}
    \end{minipage}
    \caption{
    Validation Dice coefficient and IoU per epoch for input size $160 \times 160$. 
    The values at each epoch represent the median from 30 repeated experiments. 
    The top row corresponds to cases where the number of channels in the first convolutional layer is 16, while the bottom row corresponds to cases with 64 channels.}
    \label{fig:kvasir_160}
\end{figure}

\begin{figure}[h!]
    \centering
    \begin{minipage}{0.48\textwidth}
        \centering
        \includegraphics[width=\textwidth]{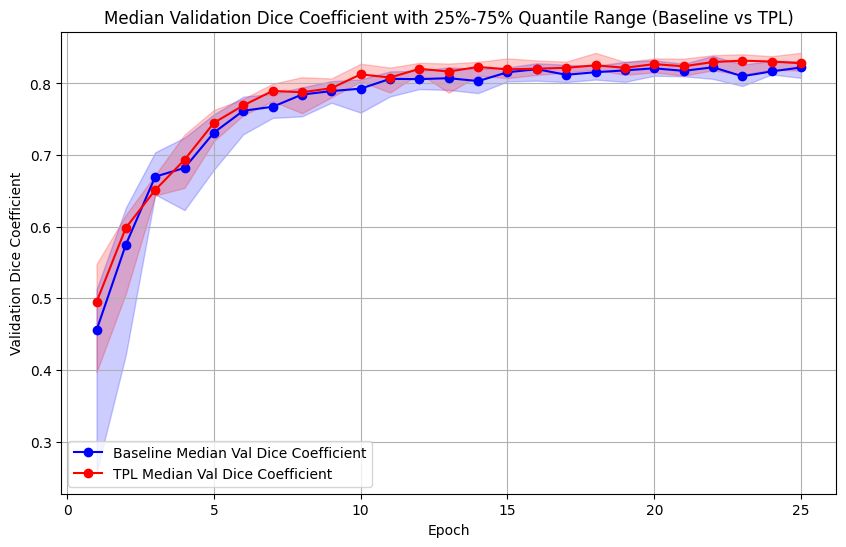}
        \subcaption{Dice score ($192 \times 192$, 16 channels)}
    \end{minipage}
    \hfill
    \begin{minipage}{0.48\textwidth}
        \centering
        \includegraphics[width=\textwidth]{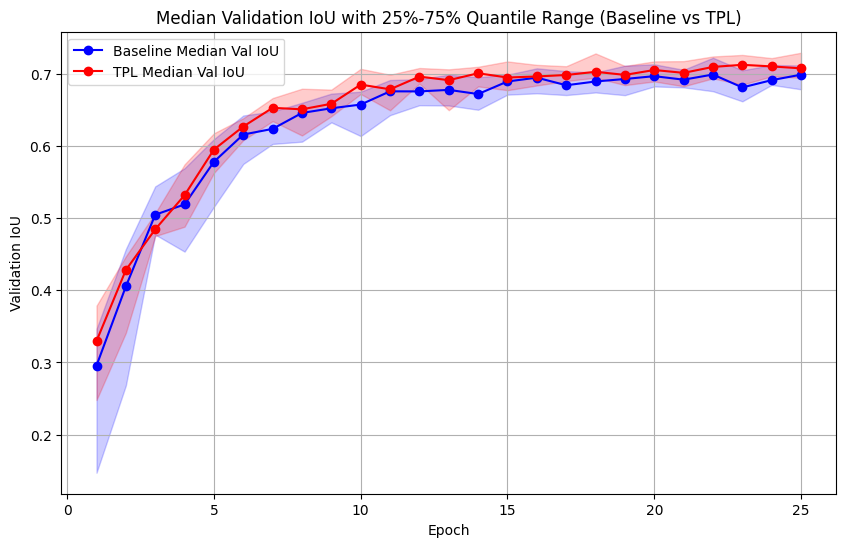}
        \subcaption{IoU score ($192 \times 192$, 16 channels)}
    \end{minipage}

    \vspace{0.3cm} %

    \begin{minipage}{0.48\textwidth}
        \centering
        \includegraphics[width=\textwidth]{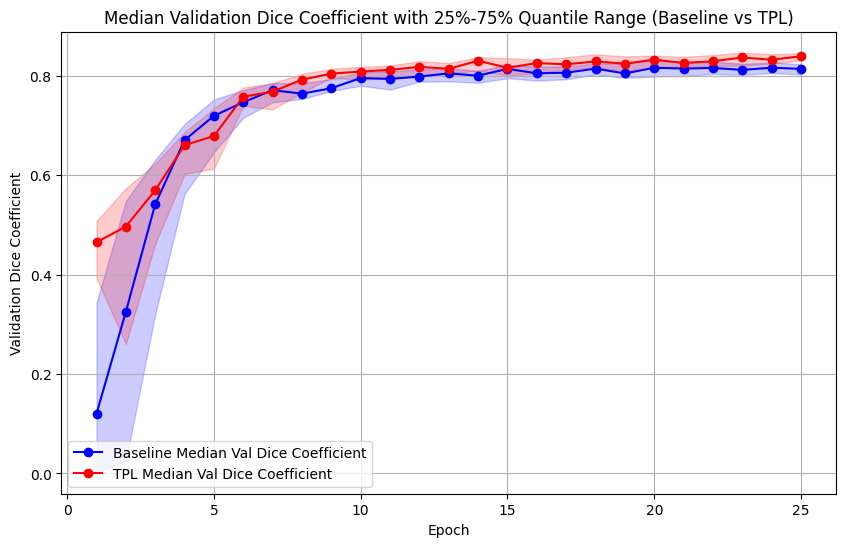}
        \subcaption{Dice score ($192 \times 192$, 64 channels)}
    \end{minipage}
    \hfill
    \begin{minipage}{0.48\textwidth}
        \centering
        \includegraphics[width=\textwidth]{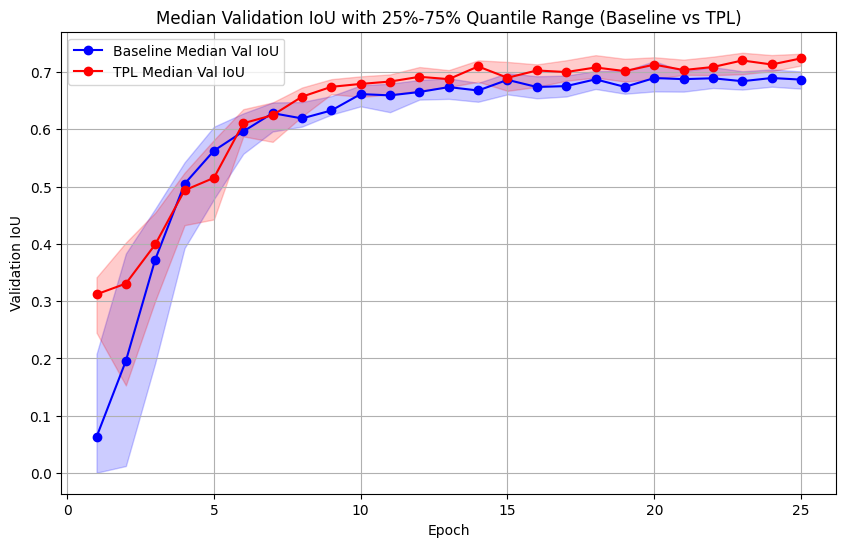}
        \subcaption{IoU score ($192 \times 192$, 64 channels)}
    \end{minipage}
    \caption{
    Validation Dice coefficient and IoU per epoch for input size $192 \times 192$. 
    The values at each epoch represent the median from 30 repeated experiments. 
    The top row corresponds to cases where the number of channels in the first convolutional layer is 16, while the bottom row corresponds to cases with 64 channels.}
    \label{fig:kvasir_192}
\end{figure}

Figures \ref{fig:kvasir_160} and \ref{fig:kvasir_192} present a comparison of the performance between the baseline model and the TPL model when the input sizes are $160 \times 160$ and $192 \times 192$, respectively. 
The two models are designed so that the input and output sizes of each layer are the same.

\clearpage

%
%
\section{Implementation}\label{appendix:implementation}

We have implemented the TensorProjection layer using Python and deep learning libraries.
The source code for the TensorProjection layer is publicly available, and you can access it at \href{https://github.com/senyuan-juncheng/TensorProjectionLayer}{https://github.com/senyuan-juncheng/TensorProjectionLayer}.

\end{appendices}

\end{document}